\pdfoutput=1

\documentclass[11pt]{article}

\usepackage{graphicx}
\usepackage{subcaption}
\usepackage{amsmath}

\usepackage[]{EACL2023}

\usepackage{times}
\usepackage{latexsym}

\usepackage[T1]{fontenc}

\usepackage[utf8]{inputenc}

\usepackage{microtype}

\usepackage{inconsolata}

\title{Transformer-based Models for Long-Form Document Matching: Challenges and Empirical Analysis}

\author{Akshita Jha \\
  Virginia Tech, Arlington, VA \\
 \texttt{akshitajha@vt.edu} \\\And
  Adithya Samavedhi \\
  Virginia Tech, Arlington, VA \\
 \texttt{adithyas@vt.edu} \\
 \And
  Vineeth Rakesh \\
  InterDigital, CA \\
 \texttt{vineethrakesh@gmail.com} \\
 \AND
  Jaideep Chandrashekar \\
  InterDigital, CA \\
 \texttt{jaideep.chandrashekar@interdigital.com} \\
 \And
  Chandan K. Reddy \\
  Virginia Tech, Arlington, VA \\
 \texttt{reddy@cs.vt.edu} \\}

\begin{document}
\maketitle
\begin{abstract}
Recent advances in the area of long document matching have primarily focused on using transformer-based models for long document encoding and matching. There are two primary challenges associated with these models. Firstly, the performance gain provided by transformer-based models comes at a steep cost – both in terms of the required training time and the resource (memory and energy) consumption. The second major limitation is their inability to handle more than a pre-defined input token length at a time. In this work, we empirically demonstrate the effectiveness of simple neural models (such as feed-forward networks, and CNNs) and simple embeddings (like GloVe, and Paragraph Vector) over transformer-based models on the task of document matching. We show that simple models outperform the more complex BERT-based models while taking significantly less training time, energy, and memory. The simple models are also more robust to variations in document length and text perturbations.
\end{abstract}

\section{Introduction}

Matching long documents (\textit{e.g.:} research papers, Wikipedia articles, patents, etc.) is an important task that can help understand the (dis)similarity between documents {for downstream tasks like long document search}. The first step towards better document matching is obtaining meaningful long document representations. Recent advances in this area have primarily focused on using transformer-based models for long document encoding and matching \cite{beltagy2020longformer, jha2022supervised, yang2020beyond, zaheer2020big}. {We use the term transformers to mean pre-trained transformers.} Despite promising results, there are two primary challenges associated with such models.
First, the performance gain provided by the huge transformer-based language models (LMs), like BERT \cite{devlin2019bert}, GPT-2 \cite{radford2019language}, and Longformer \cite{beltagy2020longformer} come at a steep cost -- both in terms of the required training time, and the resource (memory and energy) consumption. For example, the smaller $BERT_{BASE}$ model has 110 million parameters, whereas the bigger $BERT_{LARGE}$ model has a total of 340 million parameters {and fine-tuning a single $BERT_{BASE}$ model on GPU can take hours}.
The second major limitation of transformer-based models is their inability to handle more than a pre-defined input token length at a time (512 tokens for BERT, and 4096 tokens for Longformer). This is a big drawback as they cannot handle long documents like research papers, patents, long articles, etc., without using aggregation techniques \cite{reimers2019sentence}.

In this work, we empirically demonstrate that \textit{ embeddings obtained from GloVe \cite{pennington2014glove}, and Paragraph Vectors \cite{le2014distributed}} along with simple neural models, such as feed-forward networks, and CNNs, \textit{outperform several transformer-based models for the document matching task.} We define these models as simple as they take significantly less time to train and consume less memory and energy overall when compared to complex transformer-based models. {Our long document matching setting is fundamentally different from long-form question answering and sentence similarity tasks. For the latter tasks the query is `short', unlike the long document matching task where both the query and the target text are `long'}. We experiment with three different kinds of semi-structured long document datasets in English: (i) ACL Anthology research papers, (ii) English Wikipedia articles, and (iii) Patents from US Patent and Trademark Office (USPTO). Our primary contributions are summarized as follows:
\vspace{-5pt}
\begin{itemize}
    \item We provide insights into the challenges of using transformer-based models for the task of long document matching. For this task, simple neural models are as effective and take a fraction of the training time and resources to outperform transformer-based models.
    \vspace{-5pt}
    \item We provide insights into the best input embeddings for the simpler models in this task.
    \vspace{-15pt}
    \item We demonstrate that simple models are also more robust to changes in document length and text perturbation.
    \vspace{-5pt}
    \item We create benchmark long document datasets (by pre-processing ACL Anthology 2014 papers and Wikipedia articles) that will be made publicly available.
\end{itemize}

\section{Related Work}

{Early work on long document matching focused on clustering techniques \cite{friburger2002textual, huang2008similarity, strehl2000impact}}. Recently, \citet{guo2016deep} proposed a deep learning based architecture for ad-hoc retrieval when comparing documents. Some works have also used convolutional networks \cite{hu2014convolutional, pang2016text, yu2018modelling}, with weighting mechanism \cite{yang2016anmm} to generate a final query-document score. \citet{mitra2017learning} propose a combination model that uses weighted sum representation-based and interaction-based results.  \citet{yang-etal-2016-hierarchical} propose HAN, a hierarchical attention network for document matching.
\citet{jiang2019semantic} propose SMASH, a multi-depth attention based hierarchical recurrent neural network for long-document matching. However, \citet{yang2020beyond} pre-train SMITH, a transformer based hierarchical model for text matching that outperforms SMASH across multiple datasets. \citet{jha2022supervised} use supervised contrative learning for interpretable long document matching. {We compare several of these models on the required training time and resources.}
 
A growing body of literature has used transformer based model for long document encoding \cite{child2019generating, ho2019axial, kitaev2019reformer, liu-lapata-2019-hierarchical, qiu2020blockwise, yang2020html}. Longformer \cite{beltagy2020longformer} adapts transformers to use an attention mechanism that scales linearly with sequence length. Big bird \cite{zaheer2020big} uses a sparse attention mechanism that reduces BERT's quadratic dependency on the sequence length to linear. CogLTX \cite{ding2020cogltx} uses text blocks for rehearsal and decay over key sentences to overcome the insufficient long-range attentions in BERT. Transformer-XL \cite{dai2019transformer} and Compressive Tranformers \cite{rae2019compressive} compress the transformers to use attentive sequence over long text. Although promising, we demonstrate that transformer-based models are not considerably better than simple neural models on the task of long document matching.

\section{Empirical Evaluation}

Here we provide details of the simple and the transformer-based models and present an empirical comparison between them based on their overall performance, training time, resource consumption, and robustness on the document matching task.

\paragraph*{Task Formulation} We define the task of document matching as follows. Given a source document $s$, and a set of target documents $D_T$, the goal is to estimate the semantic match $\hat{y} = sim(s, t)$, where $t \in D_T$ for every document pair $(s, t)$. Similar target documents will have a higher similarity score. The document matching problem can be seen as a binary classification task, where the model predicts $1$ for similar documents, and $0$ for dissimilar documents. {We use the term ``matching" in the broad sense of document relevance (see Appendix~\ref{app:dataset})}. The models take as input a pair of documents (source and target), and compute the cosine similarity between the encoded document representations. If the cosine similarity is greater than a similarity threshold $\theta$, they are considered similar; otherwise they are considered dissimilar.


\begin{figure}
    \vspace{-20pt}
    \centering
    \includegraphics[width=0.5\linewidth]{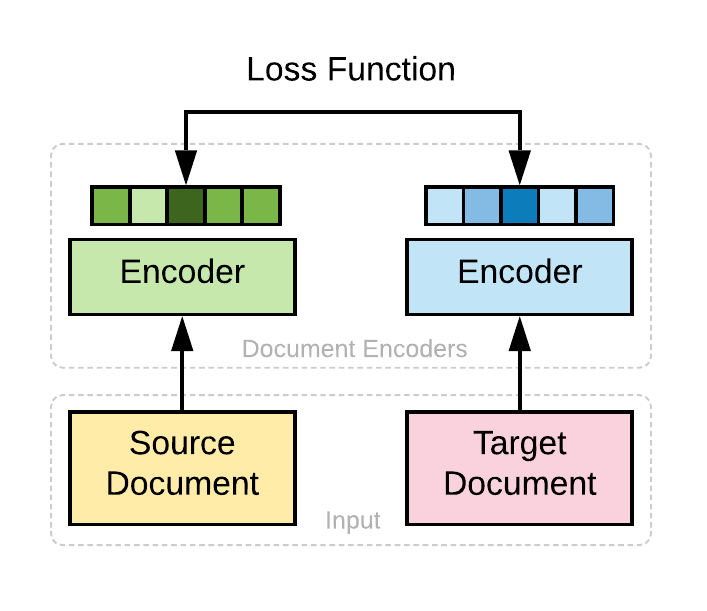}
    \vspace{-20pt}
    \caption{A schematic siamese comparison model}
    \label{fig:schematic_diagram}
    \vspace{-12pt}
\end{figure}

\begin{table*}[h]
    \small
    \centering
    \begin{tabular}{|l|c|c|c|c|c|c|c|c|c|c|c|c|}
        \hline 
        & \multicolumn{4}{c|}{\textbf{AAN}} & \multicolumn{4}{c|}{\textbf{WIKI}} & \multicolumn{4}{c|}{\textbf{PAT}} \\
        \hline
        \textbf{Model} & \textbf{P} & \textbf{R} & \textbf{F1} & \textbf{Acc} & \textbf{P} & \textbf{R} & \textbf{F1}  & \textbf{Acc} & \textbf{P} & \textbf{R} & \textbf{F1} & \textbf{Acc}\\
        \hline
        HAN-G & 0.504 & 0.881 & 0.641 & 0.607 & 0.566 & 0.584 & 0.575 & 0.775 & 0.609 & 0.848 & 0.709 & 0.522  \\
        \hline
        DSSM-T & 0.768 & 0.809 & 0.787 & {0.780} & 0.823 & 0.939 & 0.877 & 0.869 & 0.869 & 0.957 & 0.911 & 0.905 \\
        DSSM-G & 0.550 & 0.541 & 0.545 & 0.549 & 0.966 & 0.986 & 0.975 & 0.975 & 0.992 & \textbf{0.998} & \textbf{0.995} & \textbf{0.995} \\
        DSSM-D & \textbf{0.852} & 0.763 & \textbf{0.805} & \textbf{0.815} & 0.933 & 0.984 & 0.958 & 0.957 & 0.841 & 0.959 & 0.896 & 0.949 \\
        \hline
        ARC-I-G & 0.643 & 0.873 & 0.734 & 0.743 & \textbf{0.992} & \textbf{0.983} & \textbf{0.987} & \textbf{0.987} & 0.905 & 0.963 & 0.933 & 0.939 \\
        ARC-I-D & 0.841 & 0.763 & 0.800 & 0.809 & 0.987 & 0.985 & 0.986 & 0.986 & 0.967 & 0.958 & 0.962 & 0.983 \\
        \hline
        \hline
        BERT & 0.760 & \textbf{0.914} & 0.793 & 0.761 & 0.980 & 0.950 & 0.960 & 0.960 & \textbf{1.0} & 0.988 & 0.994 & 0.994\\
        LONG & 0.681 & 0.833 & 0.749 & 0.773 & 0.974 & 0.960 & 0.967 & 0.967 & \textbf{1.0} & 0.984 & 0.992 & 0.992 \\
        SMITH & 0.726 & 0.565 & 0.635 & 0.676 & 0.949 & 0.982 & 0.965 & 0.963 & 0.892 & 0.939 & 0.865 & 0.939\\
        \hline
    \end{tabular}
    \vspace{-5pt}
    \caption{Performance on the document matching task up to the model's maximum allowed input token length ($512$ for BERT; $4096$ for Longformer, $>8000$ for all other models). We experiment with Trigrams (T), GloVe (G), and Doc2Vec (D) Embeddings as input for the simple neural models. The best performance is highlighted in bold.}
    \label{tab:doc_comp}
\end{table*}

\paragraph*{Models} We pick a representative set of models from different categories and compare them by building their siamese versions (shown in Figure~\ref{fig:schematic_diagram}). The siamese network has three primary components: (i) Input (Source and Target Document), (ii) Document Encoder, and (iii) Loss Function. The source and target document encoder networks share weights. We experiment with three simple neural models: (i) DSSM: A simple feed-forward network \cite{huang2013learning}, (ii) ARC-I: A CNN-based model \cite{hu2014convolutional} , and (iii) HAN: An RNN-based Hierarchical Attention Network \cite{yang-etal-2016-hierarchical} {designed for long documents}. Their performance is compared with three state-of-the-art transformer-based models: (i) BERT \cite{devlin2019bert}, (ii) LONG: Longformer \cite{beltagy2020longformer}, and (iii) SMITH: Siamese Multi-depth Transformer based Hierarchical Encoder \cite{yang2020beyond}. {We report the mean precision, recall, F1, and accuracy over 5 folds for the best performing hyper-parameters.} The code can be found here: \url{https://github.com/AkshitaJha/SimpleModelsforLongDocumentMatching}.


\paragraph*{Datasets} {We follow the previous literature \cite{yang-etal-2016-hierarchical, jiang2019semantic, yang2020beyond} and experiment with the following three standard long document datasets}: (i) ACL Anthology Network Corpus (AAN)\footnote{\tiny{https://aan.how/download/\#aanNetworkCorpus}}, (ii) Wikipedia Articles (WIKI)\footnote{\tiny{https://dumps.wikimedia.org/enwiki/latest/enwiki-latest-pages-articles.xml.bz2}}, and (iii) USPTO Patents (PAT)\footnote{\tiny{https://github.com/google/patents-public-data}}. Each dataset consists of balanced 15,000 pairs of documents with 50\% of them being similar pairs, and the remaining being dissimilar. {The PAT dataset is an industry gold standard} but we will publicly release the pre-processed AAN and the WIKI datasets (see Appendix~\ref{app:dataset} for details). The dataset can be found here: \url{https://github.com/AkshitaJha/SimpleModelsforLongDocumentMatching}

\paragraph*{Performance on Document Matching Task}
%

We experiment with three input representations for simple neural models: (i) char-Trigram Hashing (T) \cite{huang2013learning}, (ii) GloVe Embeddings (G) \cite{pennington2014glove}, and (iii) Paragraph Vector/Doc2Vec Embeddings (D) \cite{le2014distributed} (See Appendix~\ref{tab_app:input_aggr}). 
Unlike most transformer-based models that take as input tokens up to a pre-defined length (512 for BERT, and 4096 for Longformer), simple models and SMITH have the ability to take the entire long document ($>8000$ tokens) as input. 
{Table~\ref{tab:doc_comp} demonstrates the performance of different models on the task of document matching up to their maximum allowed input document length (see Appendix~\ref{app:trans_aggr} for different document lengths.)} We observe that despite being relatively simple and not taking into account contextual embeddings, \textit{DSSM and ARC-I outperform the transformer based models using GloVe and Doc2Vec Embeddings} on the AAN, WIKI, and the PAT dataset.

\begin{figure*}[h]
\vspace{-10pt}
\begin{subfigure}[]{0.33\textwidth}
    \includegraphics[width=\textwidth]{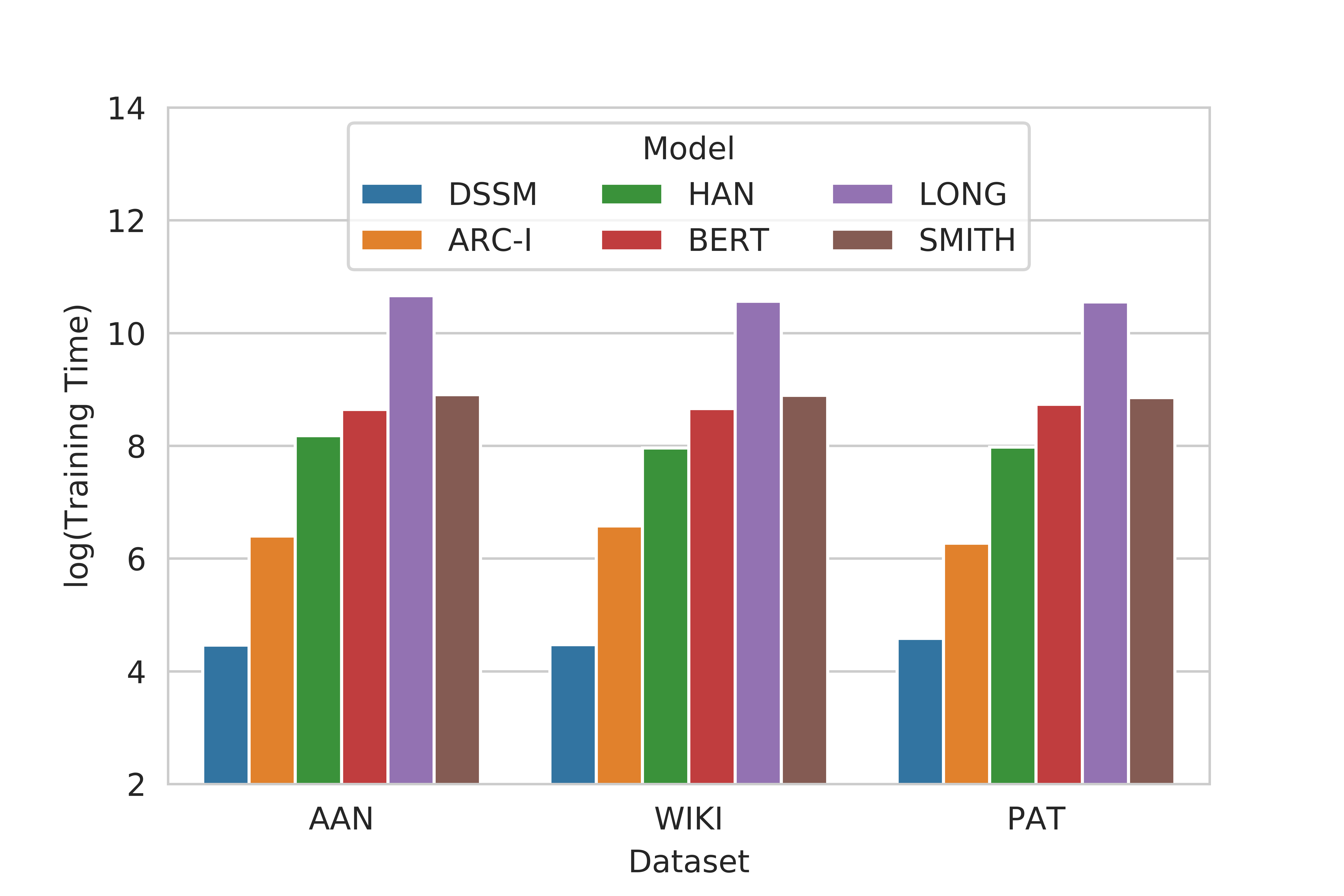}
    \caption{Training time on the \textit{log} scale}
    \label{fig:doc_vs_time}
    \end{subfigure}
    \begin{subfigure}[]{0.33\textwidth}
    \includegraphics[width=\textwidth]{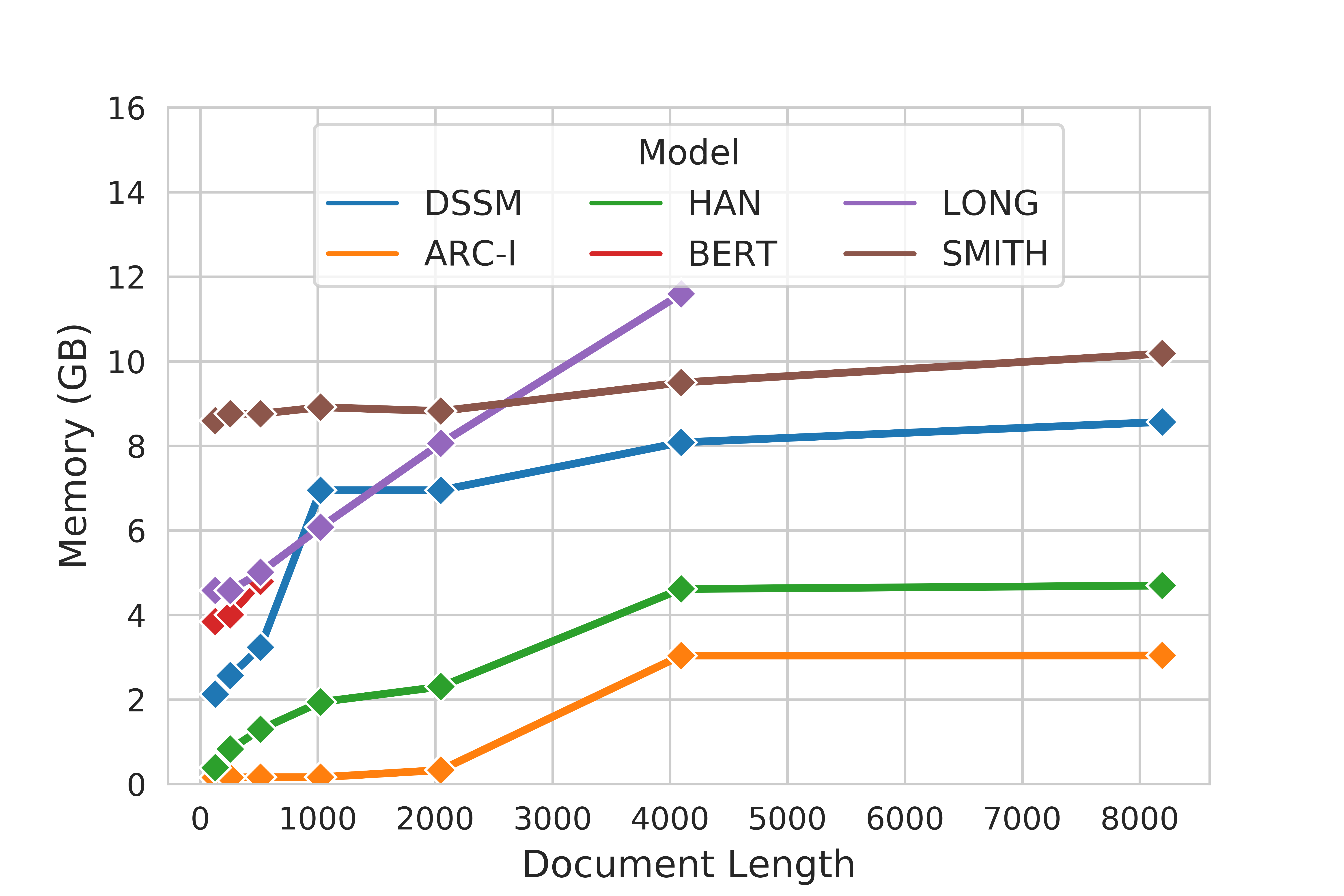}
    \caption{Memory consumption}
    \label{fig:mem_all}
    \end{subfigure}
    \begin{subfigure}[]{0.33\textwidth}
    \includegraphics[width=\textwidth]{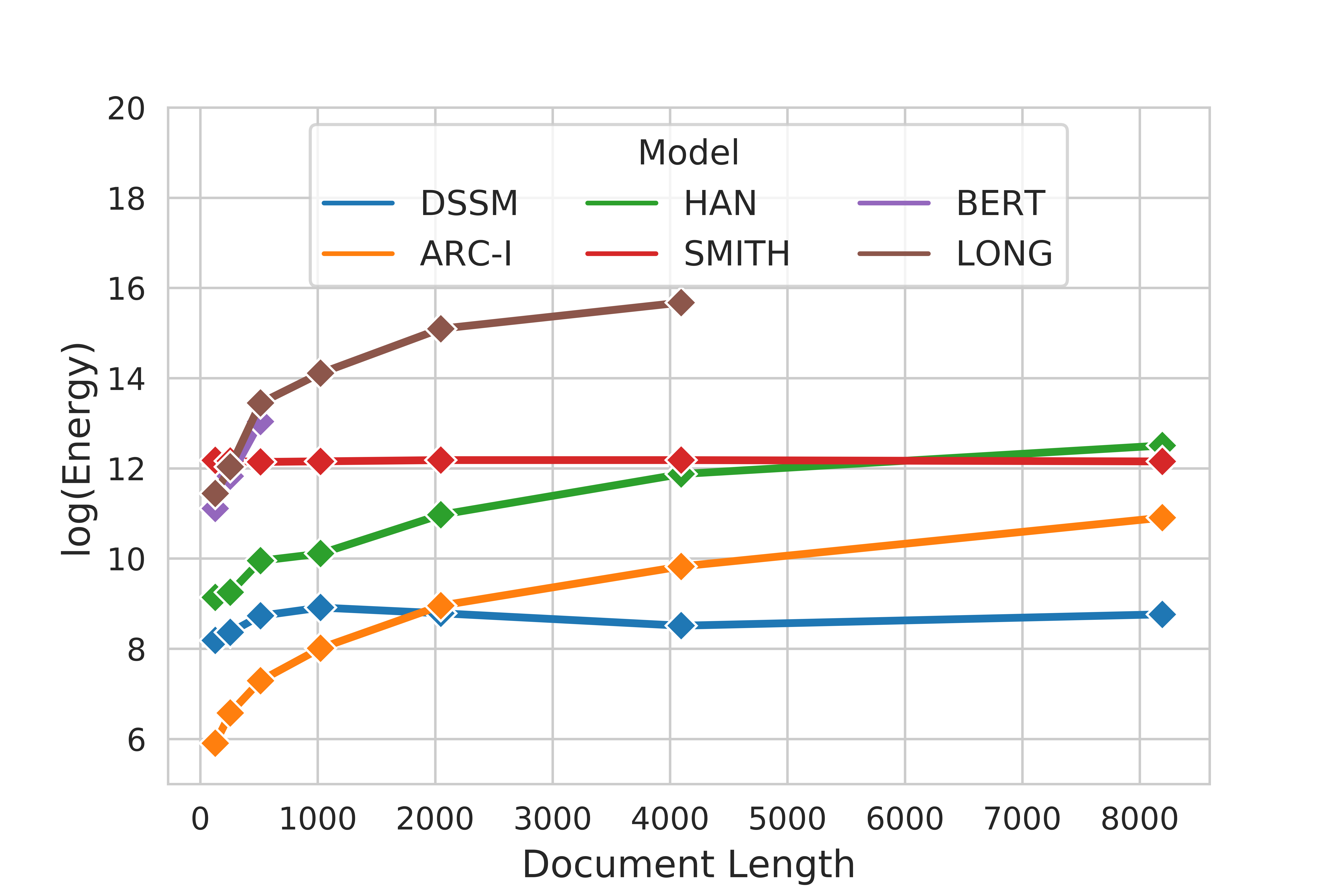}
    \caption{Energy Consumption on \textit{log} scale}
    \label{fig:energy}
    \end{subfigure}
    \vspace{-8pt}
    \caption{Comparison of simple neural models with transformer-based models based on (a) Training Time, (b) Memory Consumption, and (c) Energy Consumption.}
\end{figure*}

\begin{figure*}[h]
\vspace{-10pt}
\begin{subfigure}[]{0.33\textwidth}
    \includegraphics[width=\textwidth]{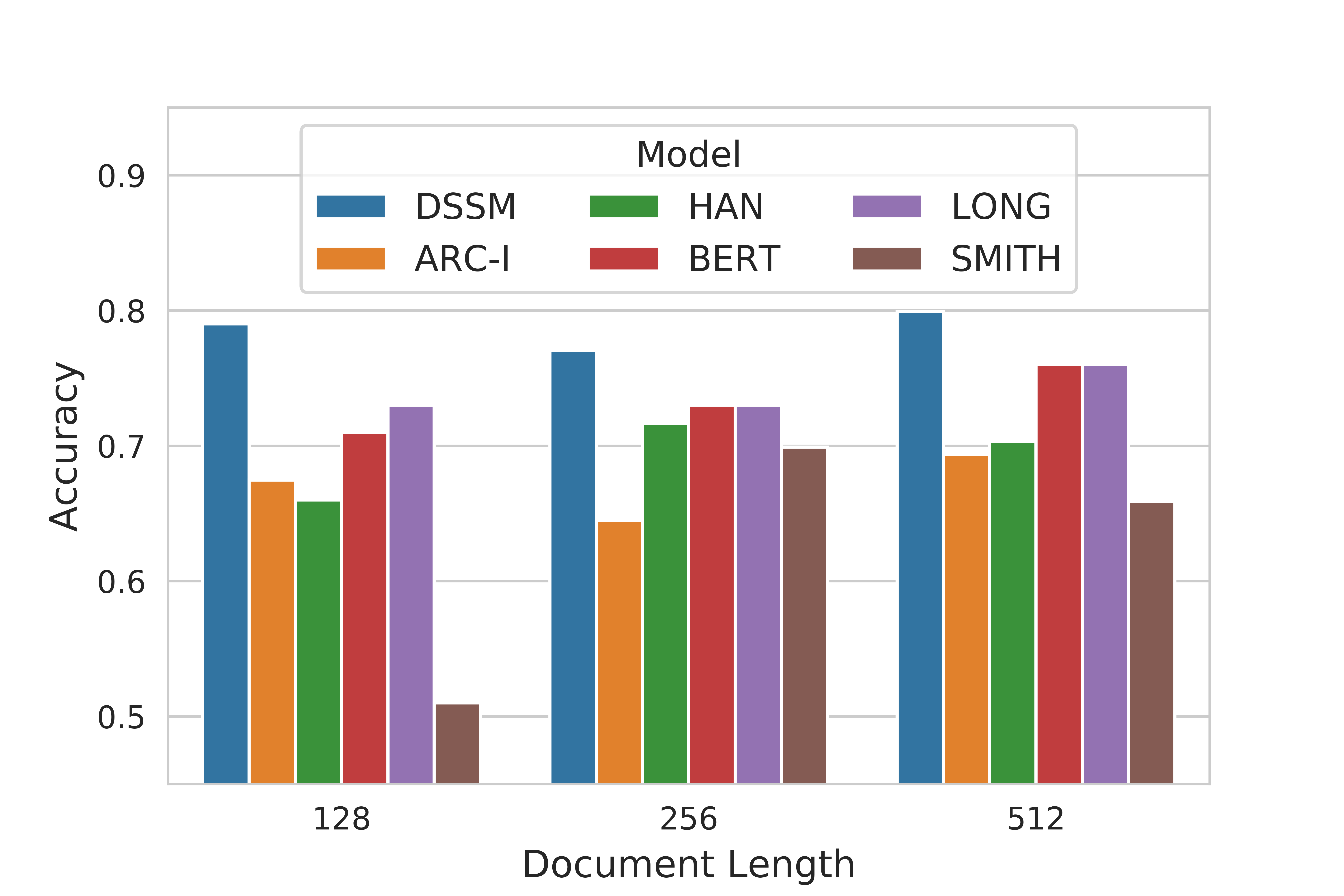}
    \caption{{Comparison with BERT [512 tokens]}}
    \label{fig:aan_bert}
    \end{subfigure}
    \begin{subfigure}[]{0.33\textwidth}
    \includegraphics[width=\textwidth]{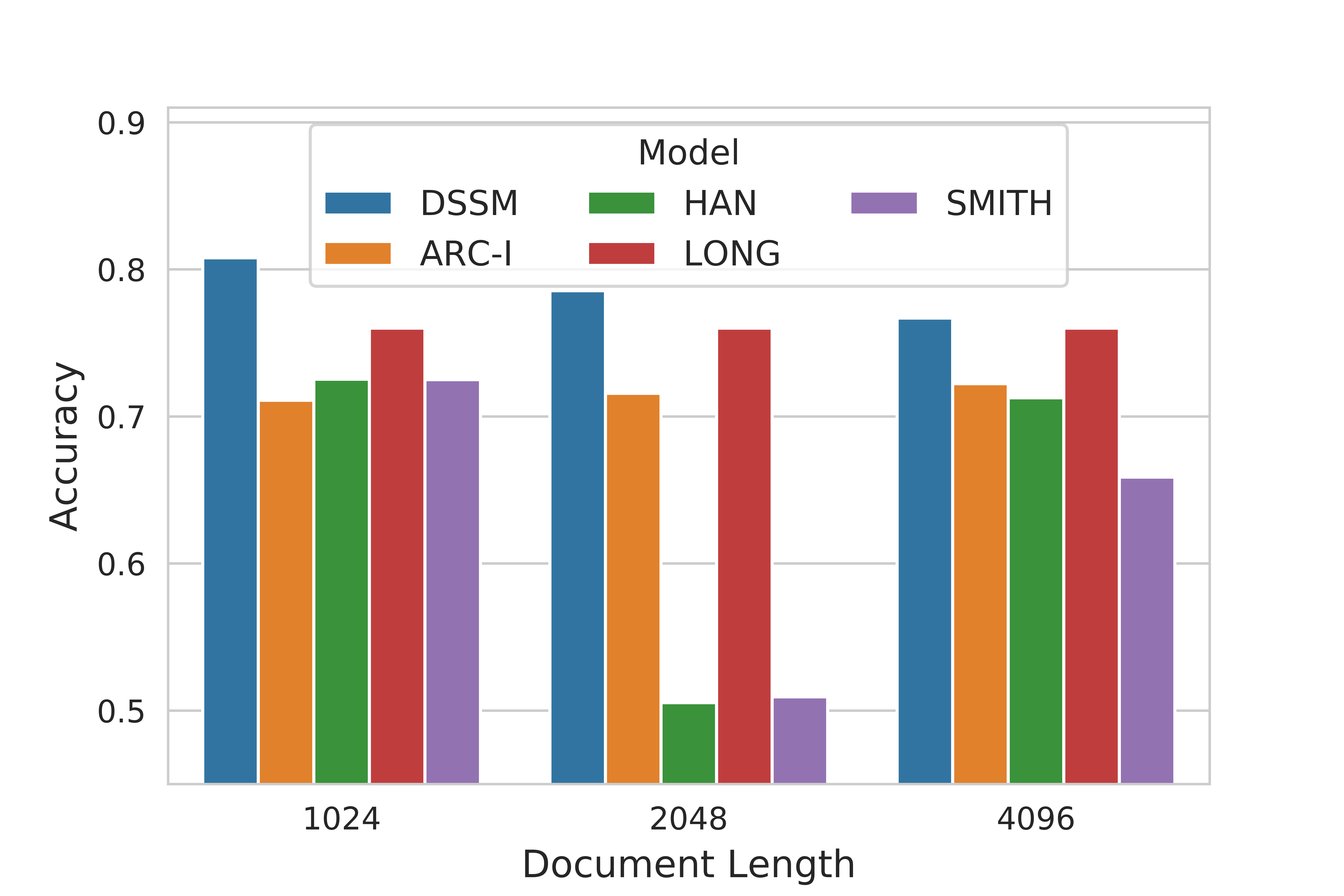}
    \caption{{Comparison with LONG [4096 tokens]}}
    \label{fig:aan_long}
    \end{subfigure}
    \begin{subfigure}[]{0.33\textwidth}
     \includegraphics[width=\textwidth]{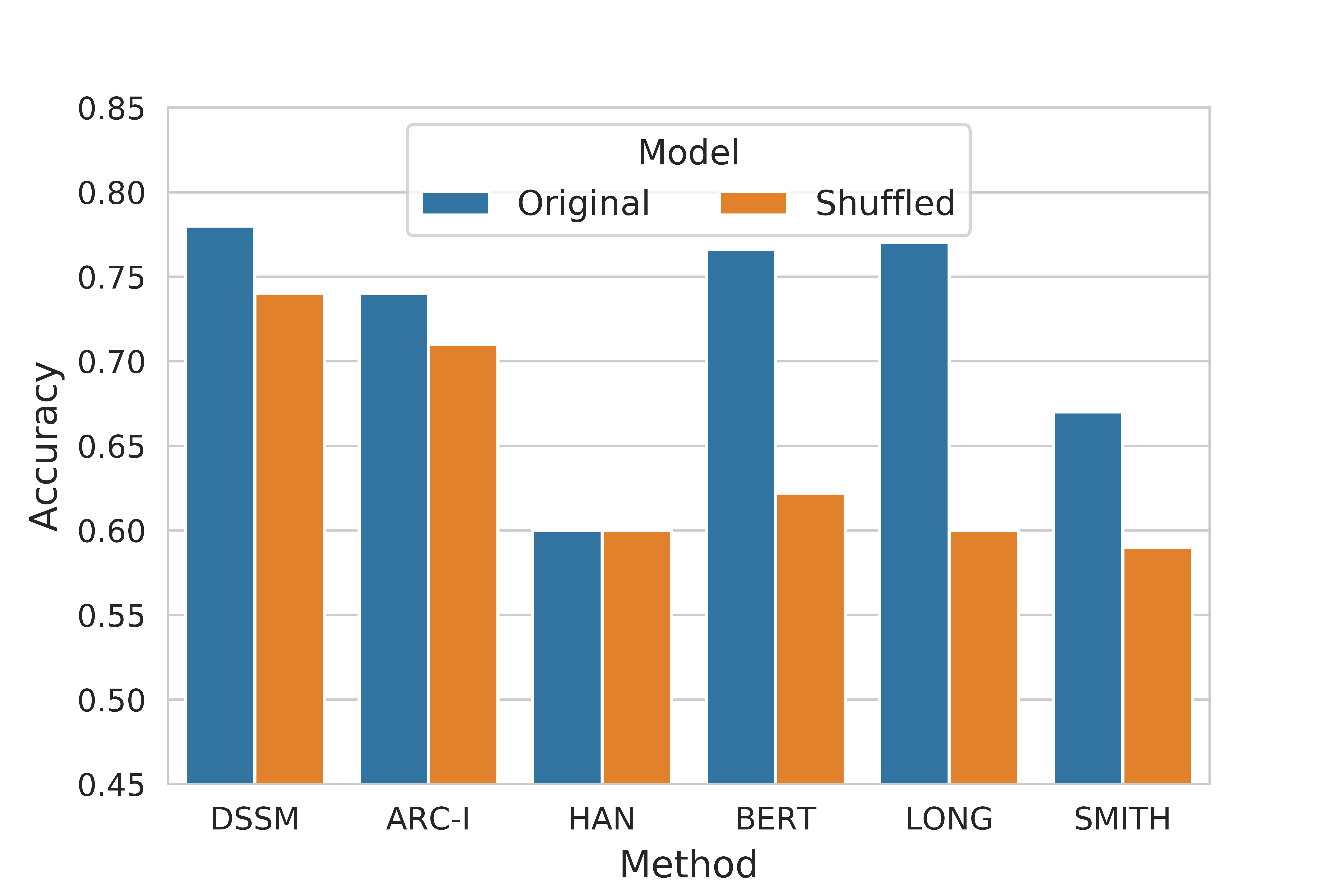}
    \caption{Original vs. Shuffled Documents}
    \label{fig:acc_textperturb}
    \end{subfigure}
    \vspace{-8pt}
    \caption{Comparison between the robustness of simple neural models and transformer-based models w.r.t. document length and text perturbation on the AAN dataset.}
    \label{fig:doclen_vs_accuracy}
\end{figure*}

\paragraph*{Training Time}
  Figure~\ref{fig:doc_vs_time} shows the training time taken to reach the best performance for every model for their maximum allowed input token lengths. {We only report the fine-tuning time after downloading the pre-trained models.} All experiments were done on a 16GiB Tesla V100. \textit{The simple models like DSSM, ARC-I, and HAN take 1/12 to 1/15 of the training time taken by the transformer-based models to outperform them on all three datasets} (see Appendix~\ref{sec_app:training_time} for the training time for different document lengths on all datasets.)

\paragraph*{Memory and Energy consumption}
 Memory consumption on a 16GiB Tesla V100, for a batch size of 1, for different models can be seen in Figure~\ref{fig:mem_all}. Compared to transformer-based models simple neural models consume significantly less memory for the same document length (12 GiB for Longformer vs. a maximum of 8 GiB for DSSM for 4000 tokens). We also compute the overall energy required for training the models to achieve their best performance (Figure~\ref{fig:energy}) by measuring the power consumption of the GPU over their training lifetime. Longformer consumes $>6 \text{MJ}$ of energy for fine-tuning on documents with 4096 tokens, BERT consumes $\approx500 \text{kJ}$ of energy for fine-tuning on documents with just 512 tokens, and the SMITH model consumes $\approx 200\text{kJ}$ of energy for fine-tuning on longer documents; whereas the simple models consume $<100\text{kJ}$ of energy for training from scratch for documents with $>8000$ tokens. 

\paragraph*{Robustness to Document Length}
We limit the maximum number of tokens in each document during training and testing, and observe the final test accuracy on the document matching task. {It should be noted that documents of different lengths are actually ‘truncated long documents’ without the complete contextual information needed to compute the actual similarity between two long documents.} Figure~\ref{fig:doclen_vs_accuracy} compares the model accuracy of simple models (with their default input embeddings) with transformer-based models upto their maximum allowed token lengths for the AAN dataset -- 512 tokens for BERT (Figure~\ref{fig:aan_bert}), and 4096 tokens for Longformer (Figure~\ref{fig:aan_long}). {DSSM outperforms the baseline models for all documents lengths. BERT and Longformer have a consistent performance on AAN for different input lengths, unlike HAN and SMITH that are not as robust to the variations in document length, although they were designed specifically for long documents.} We found similar results for WIKI and PAT dataset (see Appendix~\ref{sec_app:doclen}). {We also experiment with documents of length $>512$ tokens for BERT, and $>4096$ tokens for Longformer by aggregating the chunk representations upto their maximum allowed token length. We used the SUM and AVG aggregation techniques and observed an overall performance drop (see Table~\ref{tab_app:bert_cls_aggr}).}

\paragraph*{Robustness to Text Perturbation} {For text perturbation, we split documents into paragraphs of 512 tokens and randomly shuffle the order of these paragraphs before training different models to check for learned positional bias.} We measure their test accuracy on the original document matching task. Figure~\ref{fig:acc_textperturb} shows the model performance for all the baseline methods on AAN dataset. We observe a significant drop in the model performance for transformer-based models (BERT, Longformer, and SMITH). There is no significant change in the accuracy for the simple models -- DSSM, ARC-I, and HAN. The transformer based models are more sensitive to text perturbation. The simple models, on the other hand, use non-contextual embeddings, such as GloVE, and Doc2Vec and are more robust to text perturbation (see Appendix~\ref{sec_app:text_perturb}).


\section{Conclusion}
\vspace{-5pt}
We empirically demonstrate the trade-off of using transformer-based models for semi-structured long English documents like research papers, Wikipedia articles, and patents. {Transformer-based models have an overall strong performance and smaller variability across datasets. However, we observe that for the task of long document matching, using contextual embeddings do not provide any added advantage}. \textbf{A simple feed-forward network or a CNN-based model using GloVe or Doc2Vec embedding outperforms several state-of-the-art pre-trained transformer-based models at a fraction of their overall training time and resources (memory and energy).} These simple neural models are also more robust to changes in document length and text-perturbation.

\section{Limitation}

One of the limitations of our work is that we experimented only with long documents in English. Comparing simple neural models and transformer-based models in different languages would be an interesting study but is outside the scope of this short paper. We would also like to highlight that we use classification metrics instead of information-retrieval metrics due to the limitations of the dataset which has very few positive samples (2-3) for every document.
\bibliographystyle{acl_natbib}
\bibliography{custom}

\begin{thebibliography}{31}
\expandafter\ifx\csname natexlab\endcsname\relax\def\natexlab#1{#1}\fi

\bibitem[{Beltagy et~al.(2020)Beltagy, Peters, and
  Cohan}]{beltagy2020longformer}
Iz~Beltagy, Matthew~E Peters, and Arman Cohan. 2020.
\newblock Longformer: The long-document transformer.
\newblock \emph{arXiv preprint arXiv:2004.05150}.

\bibitem[{Child et~al.(2019)Child, Gray, Radford, and
  Sutskever}]{child2019generating}
Rewon Child, Scott Gray, Alec Radford, and Ilya Sutskever. 2019.
\newblock Generating long sequences with sparse transformers.
\newblock \emph{arXiv preprint arXiv:1904.10509}.

\bibitem[{Dai et~al.(2019)Dai, Yang, Yang, Carbonell, Le, and
  Salakhutdinov}]{dai2019transformer}
Zihang Dai, Zhilin Yang, Yiming Yang, Jaime~G Carbonell, Quoc Le, and Ruslan
  Salakhutdinov. 2019.
\newblock Transformer-xl: Attentive language models beyond a fixed-length
  context.
\newblock In \emph{Proceedings of the 57th Annual Meeting of the Association
  for Computational Linguistics}, pages 2978--2988.

\bibitem[{Devlin et~al.(2019)Devlin, Chang, Lee, and
  Toutanova}]{devlin2019bert}
Jacob Devlin, Ming-Wei Chang, Kenton Lee, and Kristina Toutanova. 2019.
\newblock Bert: Pre-training of deep bidirectional transformers for language
  understanding.
\newblock In \emph{Proceedings of the 2019 Conference of the North American
  Chapter of the Association for Computational Linguistics: Human Language
  Technologies, Volume 1 (Long and Short Papers)}, pages 4171--4186.

\bibitem[{Ding et~al.(2020)Ding, Zhou, Yang, and Tang}]{ding2020cogltx}
Ming Ding, Chang Zhou, Hongxia Yang, and Jie Tang. 2020.
\newblock Cogltx: Applying bert to long texts.
\newblock \emph{Advances in Neural Information Processing Systems}, 33.

\bibitem[{Friburger et~al.(2002)Friburger, Maurel, and
  Giacometti}]{friburger2002textual}
Nathalie Friburger, Denis Maurel, and Arnaud Giacometti. 2002.
\newblock Textual similarity based on proper names.
\newblock In \emph{Proc. of the workshop Mathematical/Formal Methods in
  Information Retrieval}, pages 155--167.

\bibitem[{Guo et~al.(2016)Guo, Fan, Ai, and Croft}]{guo2016deep}
Jiafeng Guo, Yixing Fan, Qingyao Ai, and W~Bruce Croft. 2016.
\newblock A deep relevance matching model for ad-hoc retrieval.
\newblock In \emph{Proceedings of the 25th ACM international on conference on
  information and knowledge management}, pages 55--64.

\bibitem[{Ho et~al.(2019)Ho, Kalchbrenner, Weissenborn, and
  Salimans}]{ho2019axial}
Jonathan Ho, Nal Kalchbrenner, Dirk Weissenborn, and Tim Salimans. 2019.
\newblock Axial attention in multidimensional transformers.
\newblock \emph{arXiv preprint arXiv:1912.12180}.

\bibitem[{Hu et~al.(2014)Hu, Lu, Li, and Chen}]{hu2014convolutional}
Baotian Hu, Zhengdong Lu, Hang Li, and Qingcai Chen. 2014.
\newblock Convolutional neural network architectures for matching natural
  language sentences.
\newblock \emph{Advances in neural information processing systems},
  27:2042--2050.

\bibitem[{Huang et~al.(2008)}]{huang2008similarity}
Anna Huang et~al. 2008.
\newblock Similarity measures for text document clustering.
\newblock In \emph{Proceedings of the Sixth New Zealand Computer Science
  Research Student Conference (NZCSRSC2008), Christchurch, New Zealand},
  volume~4, pages 9--56.

\bibitem[{Huang et~al.(2013)Huang, He, Gao, Deng, Acero, and
  Heck}]{huang2013learning}
Po-Sen Huang, Xiaodong He, Jianfeng Gao, Li~Deng, Alex Acero, and Larry Heck.
  2013.
\newblock Learning deep structured semantic models for web search using
  clickthrough data.
\newblock In \emph{Proceedings of the 22nd ACM international conference on
  Information \& Knowledge Management}, pages 2333--2338.

\bibitem[{Jha et~al.(2022)Jha, Rakesh, Chandrashekar, Samavedhi, and
  Reddy}]{jha2022supervised}
Akshita Jha, Vineeth Rakesh, Jaideep Chandrashekar, Adithya Samavedhi, and
  Chandan~K. Reddy. 2022.
\newblock \href {https://doi.org/10.1145/3542822} {Supervised contrastive
  learning for interpretable long-form document matching}.

\bibitem[{Jiang et~al.(2019)Jiang, Zhang, Li, Bendersky, Golbandi, and
  Najork}]{jiang2019semantic}
Jyun-Yu Jiang, Mingyang Zhang, Cheng Li, Michael Bendersky, Nadav Golbandi, and
  Marc Najork. 2019.
\newblock Semantic text matching for long-form documents.
\newblock In \emph{The World Wide Web Conference}, pages 795--806.

\bibitem[{Kitaev et~al.(2019)Kitaev, Kaiser, and Levskaya}]{kitaev2019reformer}
Nikita Kitaev, Lukasz Kaiser, and Anselm Levskaya. 2019.
\newblock Reformer: The efficient transformer.
\newblock In \emph{International Conference on Learning Representations}.

\bibitem[{Le and Mikolov(2014)}]{le2014distributed}
Quoc Le and Tomas Mikolov. 2014.
\newblock Distributed representations of sentences and documents.
\newblock In \emph{International conference on machine learning}, pages
  1188--1196. PMLR.

\bibitem[{Liu and Lapata(2019)}]{liu-lapata-2019-hierarchical}
Yang Liu and Mirella Lapata. 2019.
\newblock \href {https://doi.org/10.18653/v1/P19-1500} {Hierarchical
  transformers for multi-document summarization}.
\newblock In \emph{Proceedings of the 57th Annual Meeting of the Association
  for Computational Linguistics}, pages 5070--5081, Florence, Italy.
  Association for Computational Linguistics.

\bibitem[{Mitra et~al.(2017)Mitra, Diaz, and Craswell}]{mitra2017learning}
Bhaskar Mitra, Fernando Diaz, and Nick Craswell. 2017.
\newblock Learning to match using local and distributed representations of text
  for web search.
\newblock In \emph{Proceedings of the 26th International Conference on World
  Wide Web}, pages 1291--1299.

\bibitem[{Pang et~al.(2016)Pang, Lan, Guo, Xu, Wan, and Cheng}]{pang2016text}
Liang Pang, Yanyan Lan, Jiafeng Guo, Jun Xu, Shengxian Wan, and Xueqi Cheng.
  2016.
\newblock Text matching as image recognition.
\newblock In \emph{Proceedings of the AAAI Conference on Artificial
  Intelligence}, volume~30.

\bibitem[{Pennington et~al.(2014)Pennington, Socher, and
  Manning}]{pennington2014glove}
Jeffrey Pennington, Richard Socher, and Christopher~D Manning. 2014.
\newblock Glove: Global vectors for word representation.
\newblock In \emph{Proceedings of the 2014 conference on empirical methods in
  natural language processing (EMNLP)}, pages 1532--1543.

\bibitem[{Qiu et~al.(2020)Qiu, Ma, Levy, Yih, Wang, and
  Tang}]{qiu2020blockwise}
Jiezhong Qiu, Hao Ma, Omer Levy, Wen-tau Yih, Sinong Wang, and Jie Tang. 2020.
\newblock Blockwise self-attention for long document understanding.
\newblock In \emph{Proceedings of the 2020 Conference on Empirical Methods in
  Natural Language Processing: Findings}, pages 2555--2565.

\bibitem[{Radev et~al.(2013)Radev, Muthukrishnan, Qazvinian, and
  Abu-Jbara}]{aan}
Dragomir~R. Radev, Pradeep Muthukrishnan, Vahed Qazvinian, and Amjad Abu-Jbara.
  2013.
\newblock \href {https://doi.org/10.1007/s10579-012-9211-2} {The acl anthology
  network corpus}.
\newblock \emph{Language Resources and Evaluation}, pages 1--26.

\bibitem[{Radford et~al.(2019)Radford, Wu, Child, Luan, Amodei, and
  Sutskever}]{radford2019language}
Alec Radford, Jeffrey Wu, Rewon Child, David Luan, Dario Amodei, and Ilya
  Sutskever. 2019.
\newblock Language models are unsupervised multitask learners.
\newblock \emph{OpenAI blog}, 1(8):9.

\bibitem[{Rae et~al.(2019)Rae, Potapenko, Jayakumar, Hillier, and
  Lillicrap}]{rae2019compressive}
Jack~W Rae, Anna Potapenko, Siddhant~M Jayakumar, Chloe Hillier, and Timothy~P
  Lillicrap. 2019.
\newblock Compressive transformers for long-range sequence modelling.
\newblock In \emph{International Conference on Learning Representations}.

\bibitem[{Reimers and Gurevych(2019)}]{reimers2019sentence}
Nils Reimers and Iryna Gurevych. 2019.
\newblock Sentence-bert: Sentence embeddings using siamese bert-networks.
\newblock In \emph{Proceedings of the 2019 Conference on Empirical Methods in
  Natural Language Processing and the 9th International Joint Conference on
  Natural Language Processing (EMNLP-IJCNLP)}, pages 3973--3983.

\bibitem[{Strehl et~al.(2000)Strehl, Ghosh, and Mooney}]{strehl2000impact}
Alexander Strehl, Joydeep Ghosh, and Raymond Mooney. 2000.
\newblock Impact of similarity measures on web-page clustering.
\newblock In \emph{Workshop on artificial intelligence for web search (AAAI
  2000)}, volume~58, page~64.

\bibitem[{Yang et~al.(2020{\natexlab{a}})Yang, Ng, Smyth, and
  Dong}]{yang2020html}
Linyi Yang, Tin Lok~James Ng, Barry Smyth, and Riuhai Dong. 2020{\natexlab{a}}.
\newblock Html: Hierarchical transformer-based multi-task learning for
  volatility prediction.
\newblock In \emph{Proceedings of The Web Conference 2020}, pages 441--451.

\bibitem[{Yang et~al.(2016{\natexlab{a}})Yang, Ai, Guo, and
  Croft}]{yang2016anmm}
Liu Yang, Qingyao Ai, Jiafeng Guo, and W~Bruce Croft. 2016{\natexlab{a}}.
\newblock anmm: Ranking short answer texts with attention-based neural matching
  model.
\newblock In \emph{Proceedings of the 25th ACM international on conference on
  information and knowledge management}, pages 287--296.

\bibitem[{Yang et~al.(2020{\natexlab{b}})Yang, Zhang, Li, Bendersky, and
  Najork}]{yang2020beyond}
Liu Yang, Mingyang Zhang, Cheng Li, Michael Bendersky, and Marc Najork.
  2020{\natexlab{b}}.
\newblock Beyond 512 tokens: Siamese multi-depth transformer-based hierarchical
  encoder for long-form document matching.
\newblock In \emph{Proceedings of the 29th ACM International Conference on
  Information \& Knowledge Management}, pages 1725--1734.

\bibitem[{Yang et~al.(2016{\natexlab{b}})Yang, Yang, Dyer, He, Smola, and
  Hovy}]{yang-etal-2016-hierarchical}
Zichao Yang, Diyi Yang, Chris Dyer, Xiaodong He, Alex Smola, and Eduard Hovy.
  2016{\natexlab{b}}.
\newblock \href {https://doi.org/10.18653/v1/N16-1174} {Hierarchical attention
  networks for document classification}.
\newblock In \emph{Proceedings of the 2016 Conference of the North {A}merican
  Chapter of the Association for Computational Linguistics: Human Language
  Technologies}, pages 1480--1489, San Diego, California. Association for
  Computational Linguistics.

\bibitem[{Yu et~al.(2018)Yu, Qiu, Jiang, Huang, Song, Chu, and
  Chen}]{yu2018modelling}
Jianfei Yu, Minghui Qiu, Jing Jiang, Jun Huang, Shuangyong Song, Wei Chu, and
  Haiqing Chen. 2018.
\newblock Modelling domain relationships for transfer learning on
  retrieval-based question answering systems in e-commerce.
\newblock In \emph{Proceedings of the Eleventh ACM International Conference on
  Web Search and Data Mining}, pages 682--690.

\bibitem[{Zaheer et~al.(2020)Zaheer, Guruganesh, Dubey, Ainslie, Alberti,
  Ontanon, Pham, Ravula, Wang, Yang et~al.}]{zaheer2020big}
Manzil Zaheer, Guru Guruganesh, Avinava Dubey, Joshua Ainslie, Chris Alberti,
  Santiago Ontanon, Philip Pham, Anirudh Ravula, Qifan Wang, Li~Yang, et~al.
  2020.
\newblock Big bird: Transformers for longer sequences.
\newblock \emph{arXiv preprint arXiv:2007.14062}.

\end{thebibliography}

\newpage
\clearpage
\appendix
\section{Appendix}
\label{sec:appendix}
\subsection{Model Description}

The encoder networks and their default inputs have been described below:

\subsubsection{Simple Models}\label{app:simple_models}
\begin{itemize}
    \item \textbf{DSSM} \cite{huang2013learning}: A \textit{simple three-layered feed forward network} that takes as input the vectorized representation of a document. 

    \item \textbf{ARC-I} \cite{hu2014convolutional}: A \textit{CNN-based model} that takes as input a 2D-matrix representation of a document where words in the sentences are represented using \textit{GloVe embeddings} \cite{pennington2014glove}.  These are then passed through convolutional and max-pooling layers to finally obtain a document representation for both source and target documents, independently. The document representations are concatenated and passed through a multi-layer perceptron to predict if the pair of documents are similar or not. 

    \item \textbf{Hierarchical Attention Network (HAN)} \cite{yang-etal-2016-hierarchical}: A hierarchical \textit{GRU-based model} with attention mechanism that aggregates \textit{GloVe embeddings} at word level into sentence representations to arrive at the final document representation. 
\end{itemize}

\subsubsection{Transformer-Based Models}
\begin{itemize}
    \item \textbf{BERT} \cite{devlin2019bert}: A siamese matching model with \textit{BERT}. For long document inputs, BERT only uses the first 512 tokens of each document. We use a pre-trained $BERT_{BASE}$ model which is fine-tuned during training. 
    
    \item \textbf{Longformers (LONG)} \cite{beltagy2020longformer} : A siamese model with transformer-based \textit{Longformers} for long sequences. It has an attention mechanism that scales linearly with sequence length and takes as input a maximum of 4096 tokens. We consider an attention window of size 256.
    
    \item \textbf{Siamese Multi-depth Transformer based Hierarchical Encoder (SMITH)} \cite{yang2020beyond}:
    A \textit{transformer-based hierarchical encoder} which is the current SOTA model for long-form document matching task. 
\end{itemize}

\subsubsection{Implementation Details}
{For all the models presented in the paper, we use the same architecture as the original papers. We tune the hyperparamters and report the best results. The DSSM, ARC-I, HAN, and SMITH models were implemented in Tensorflow. BERT and Longformer were implemented in PyTorch. DSSM has hidden units of dimension 300 for its hidden layers and an output dimension of 128. The learning rate was 0.0075. ARC-I takes as input a 2D matrix of the size [no. of sentences x sentence length]. This is given as input to two 1D-convolutional (filter size of 200, kernel size 3) and MaxPooling layers of size 2, in order to get the final document representations. The representations of both the source and the target documents are concatenated and passed through a two-layer multi-layer perceptron. The first hidden layer of the MLP has a dimension of 64 with ReLU activation. The second layer has 1 node and sigmoid activation which predicts if the pair of documents are similar or not. The learning rate was set to 0.00075. HAN uses a bi-Directional GRU layer and applies attention mechanism to arrive at final sentence representation for the source and the target documents with a learning rate of 0.001. We use the pre-trained $\text{BERT}_\text{BASE}$\footnote{https://huggingface.co/transformers/model\_doc/bert.html} and the Longformer \footnote{https://huggingface.co/transformers/model\_doc/longformer.html} models provided by the Huggingface library. The SMITH code was publicly available. The BERT, Longformer, and SMITH models are fine-tuned during training. All other models are trained from scratch. The learning rate is set to 5e-5 for the transformer based models. We use an Adam optimizer for all models with a weight decay of 0.01. We use binary cross entropy as the loss function for the simple models, pairwise loss for BERT and Longformer, and triplet loss for SMITH. These loss functions resulted in the best performance for the model. The three datasets are split into 80-10-10 for train, validation, and test sets, respectively. We use cosine similarity and the similarity threshold $\theta$ is set to 0.5. We perform 5 fold cross-validation and use early stopping on the validation set to prevent over-fitting. The models were trained on one 16GB Tesla V100 GPU.}

\subsection{Dataset} \label{app:dataset}

\begin{itemize}
    \item \textbf{ACL Anthology Network Corpus (AAN)\footnote{https://aan.how/download/\#aanNetworkCorpus}}: The AAN corpus \cite{aan} consists of 23,766 papers written by 18,862 authors in 373 venues related to NLP and forms a citation network. Each paper is represented by a node with directed edges connecting a paper (the parent node) to all its cited papers (children nodes). Papers that have been cited by the parent paper are treated as similar samples \cite{jiang2019semantic}. For every similar sample, an irrelevant paper is randomly chosen to create a balanced dataset. Sets of similar papers are given the same labels. To prevent leakage of information and make the task more difficult, the references and the abstract sections are removed. Papers without any content are also removed. {We then randomly sample 15,000 research paper pairs for our experiment.}
    
    \item \textbf{Wikipedia (WIKI)}\footnote{https://dumps.wikimedia.org/enwiki/latest/enwiki-latest-pages-articles.xml.bz2}: We use the Wikipedia dump, and adopt a similar methodology proposed by Jiang \textit{et al.} \cite{jiang2019semantic} to process this data. {From the Wikipedia dump containing ~6 million articles, we randomly sampled 250,000 articles along with the articles present in their outlinks.} We create a dataset of similar Wikipedia articles by assuming that similar articles have similar outgoing links. The Jaccard similarity between the outgoing links of the source and the target articles is calculated. If the Jaccard similarity $> 0.5$, the documents are assumed to be similar, otherwise they are considered dissimilar. Only articles with two or more similar articles are selected. {We then randomly sample 15,000 research paper pairs for our experiment.}
   
    \item \textbf{Patent (PAT)}\footnote{https://github.com/google/patents-public-data}: {The patent dataset is an internally curated industry gold-standard}. {This dataset consists of patents sampled from the publicly available USPTO patents {belonging to four different categories: video, wireless, image compression, and network compression.} A patent document is extremely long and primarily consists of (i) Abstract, (ii) Claims, and (iii) Description sections. We only make use of the Claims and the Description sections for our experiments {to prevent leakage of information from Abstracts. Three internal human annotators, with expert domain knowledge, were given pairs of documents and were asked to label them as similar or dissimilar based based on the technology presented in the patents.} They referred to the Abstract, Claims, and CPC} Codes\footnote{\tiny{https://www.uspto.gov/web/patents/classification/cpc/html/cpc.html}} of the patents to measure the similarity. The final document content similarity label was based on majority vote.
\end{itemize}

The dataset statistics can be found in Table~\ref{tab_app:dataset_stats}. {We would like to note that although considering only the cited papers and outgoing links for AAN and Wikipedia articles, respectively is not the most optimal approach for creating similar document pairs, we adopt it for the following two reasons: (i) We do not have annotated fine-grained similarity scores for AAN and WIKI datasets, and (ii) We follow an approach similar to the previously published work \cite{jiang2019semantic, yang-etal-2016-hierarchical, yang2020beyond}. We use the term ``document matching" or ``document similarity" in the broad sense of ``citation matching" or ``document relevance" -- Given a pair of documents, are the two documents relevant to each other and should they be cited by each other.}

\begin{table*}[!h]
    \centering
    \begin{tabular}{|l|c|c|c|c|c|}
        \hline 
        \textbf{Dataset} & \textbf{Avg \#Tokens} & \textbf{Max \#Tokens} & \textbf{Avg \#Sentences} & \textbf{Max \#Sentences} & \textbf{Vocabulary} \\
        \hline
        AAN  & 5,381.1 & 54,556 & 215.7 & 2,183 & 515,422  \\
        WIKI & 3,777.0 & 26,172 & 190.6 & 1,685 & 1,151,309  \\
        PAT & 8,177.4 & 50,322 & 214.1 & 2,709 & 220,023 \\
        \hline
    \end{tabular}
    \caption{Dataset statistics}
    \label{tab_app:dataset_stats}
\end{table*}

\subsection{Performance on Document Matching Task}

{Table~\ref{tab:avg_doc_comp} shows the average performance of simple models when compared to transformer based models on the document matching task for AAN, WIKI, and PAT datasets. ARC-I with Doc2Vec embeddings has the best average precision and accuracy. The average F1 score is comparable to BERT which re-emphasizes the benefits of using simple models for document matching.}

\begin{table*}[!h]
    \centering
    \begin{tabular}{|l|c|c|c|c|}
        \hline
        \textbf{Model} & \textbf{P} & \textbf{R} & \textbf{F1} & \textbf{Acc} \\
        \hline
        HAN-G & 0.559 & 0.771 & 0.641 & 0.634 \\
        \hline
        DSSM-T & 0.820 & 0.901 & 0.858 & {0.851}  \\
        DSSM-G & 0.836 & 0.841 & 0.838 & 0.839 \\
        DSSM-D & {0.875} & 0.902 & {0.886} & {0.907} \\
        \hline
        ARC-I-G & 0.846 & 0.939 & 0.884 & 0.889 \\
        ARC-I-D & \textbf{0.931} & 0.902 & \textbf{0.916} & \textbf{0.926} \\
        \hline
        \hline
        BERT & 0.913 & \textbf{0.950} & 0.915 & 0.905 \\
        LONG & 0.885 & 0.925 & 0.902 & 0.910  \\
        SMITH & 0.855 & 0.828 & 0.821 & 0.860 \\
        \hline
    \end{tabular}
    \vspace{-5pt}
    \caption{Average performance across AAN, WIKI, and PAT datasets on the document matching task (shown in Table 1). We experiment with Trigrams (T), GloVe (G), and Doc2Vec (D) Embeddings as input for the simple neural models. The best performance is highlighted in bold.}
    \label{tab:avg_doc_comp}
\end{table*}

\subsection{Training Time for Different Document Lengths}\label{sec_app:training_time}
\begin{figure*}[h!]
\begin{subfigure}[]{0.33\linewidth}
    \includegraphics[width=\textwidth]{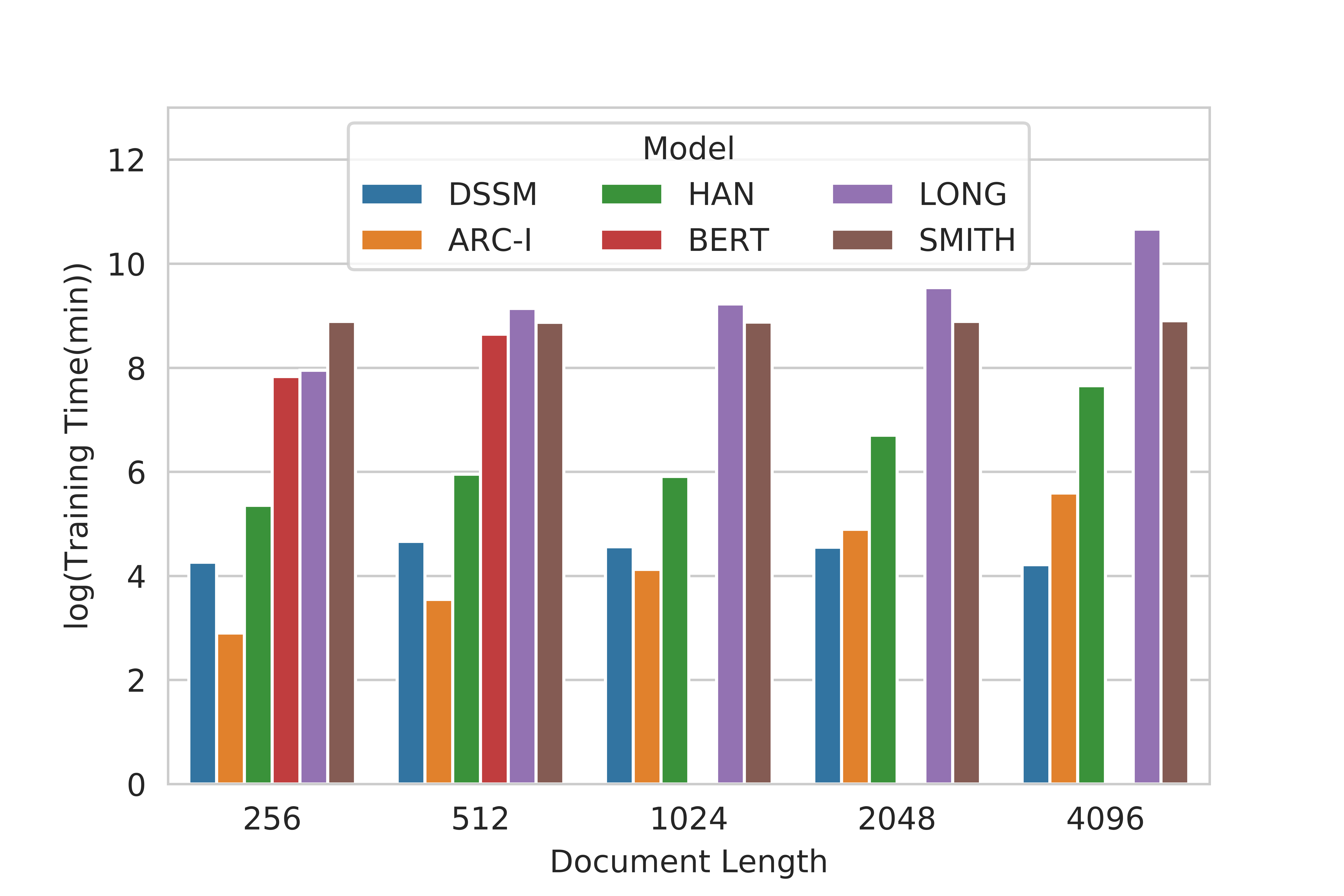}
    \caption{AAN}
    \end{subfigure}
    \begin{subfigure}[]{0.33\linewidth}
    \includegraphics[width=\textwidth]{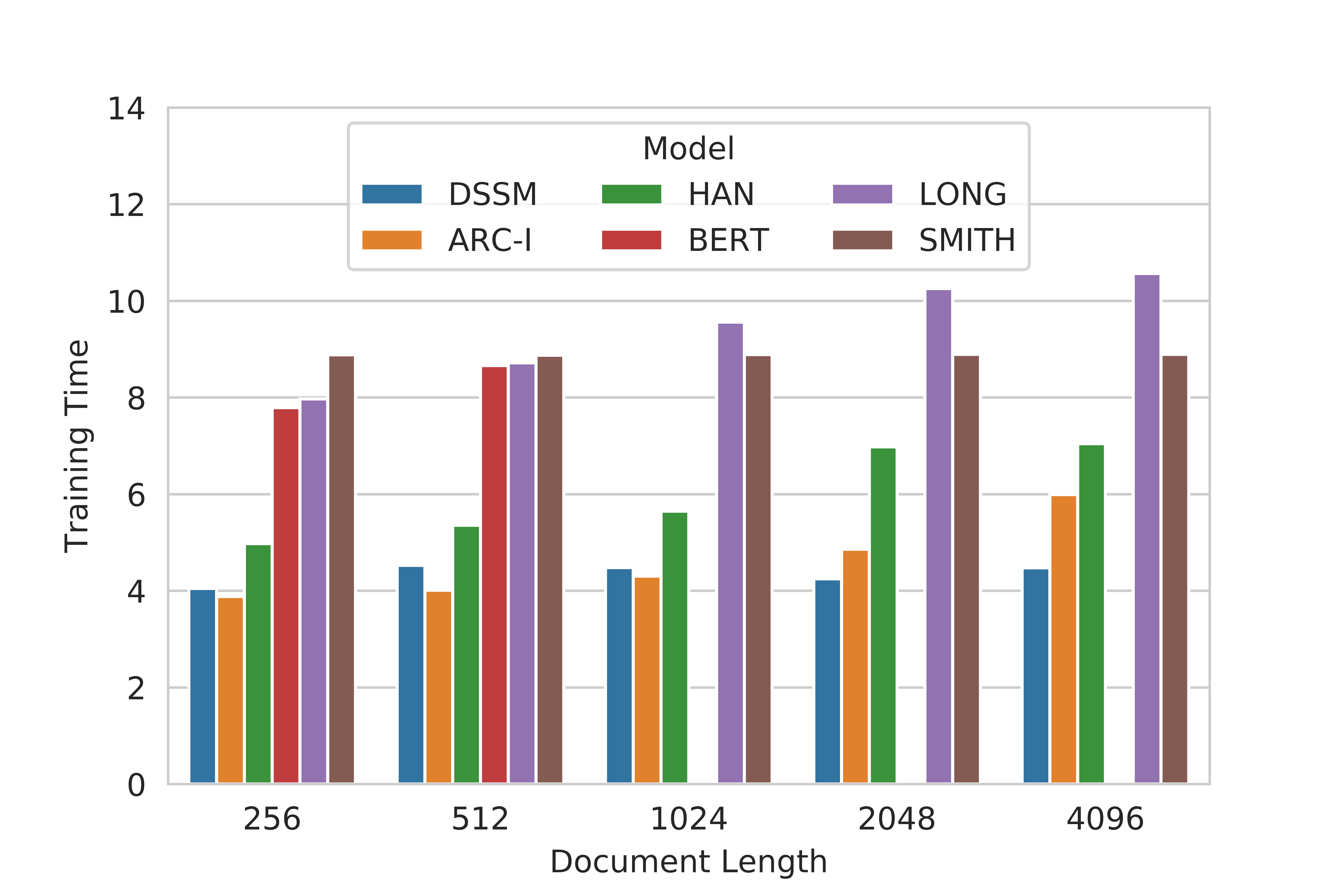}
    \caption{WIKI}
    \end{subfigure}
    \begin{subfigure}[]{0.33\linewidth}
    \includegraphics[width=\textwidth]{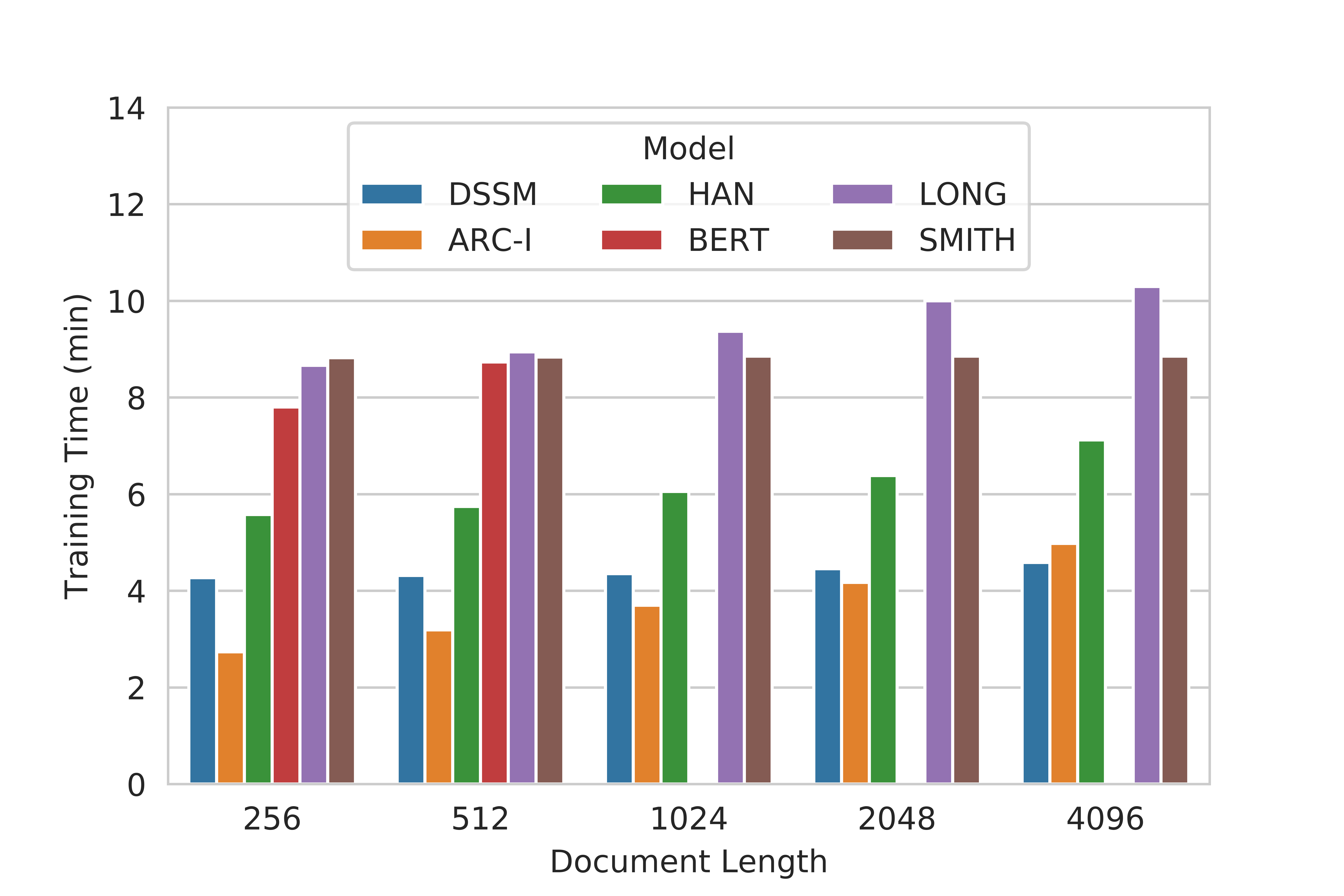}
    \caption{PAT}
    \end{subfigure}
    \caption{Doc Length vs Training Time (in \textit{log} scale) for different document lengths.}
    \label{fig_app:doclen_vs_time}
\end{figure*}

The training time for different document lengths for all three datasets can be seen in Figure~\ref{fig_app:doclen_vs_time}. BERT can only take up to 512 tokens. DSSM, ARC-1, and HAN take considerably less time to train when compared to transformer-based.

\subsection{Robustness to Document Length}\label{sec_app:doclen}

\begin{figure*}[h!]
\begin{subfigure}[]{0.33\linewidth}
    \includegraphics[width=\textwidth]{imgs/doclen_vs_acc/aan_doclen_vs_acc_bert.png}
    \caption{AAN}
    \end{subfigure}
    \begin{subfigure}[]{0.33\linewidth}
    \includegraphics[width=\textwidth]{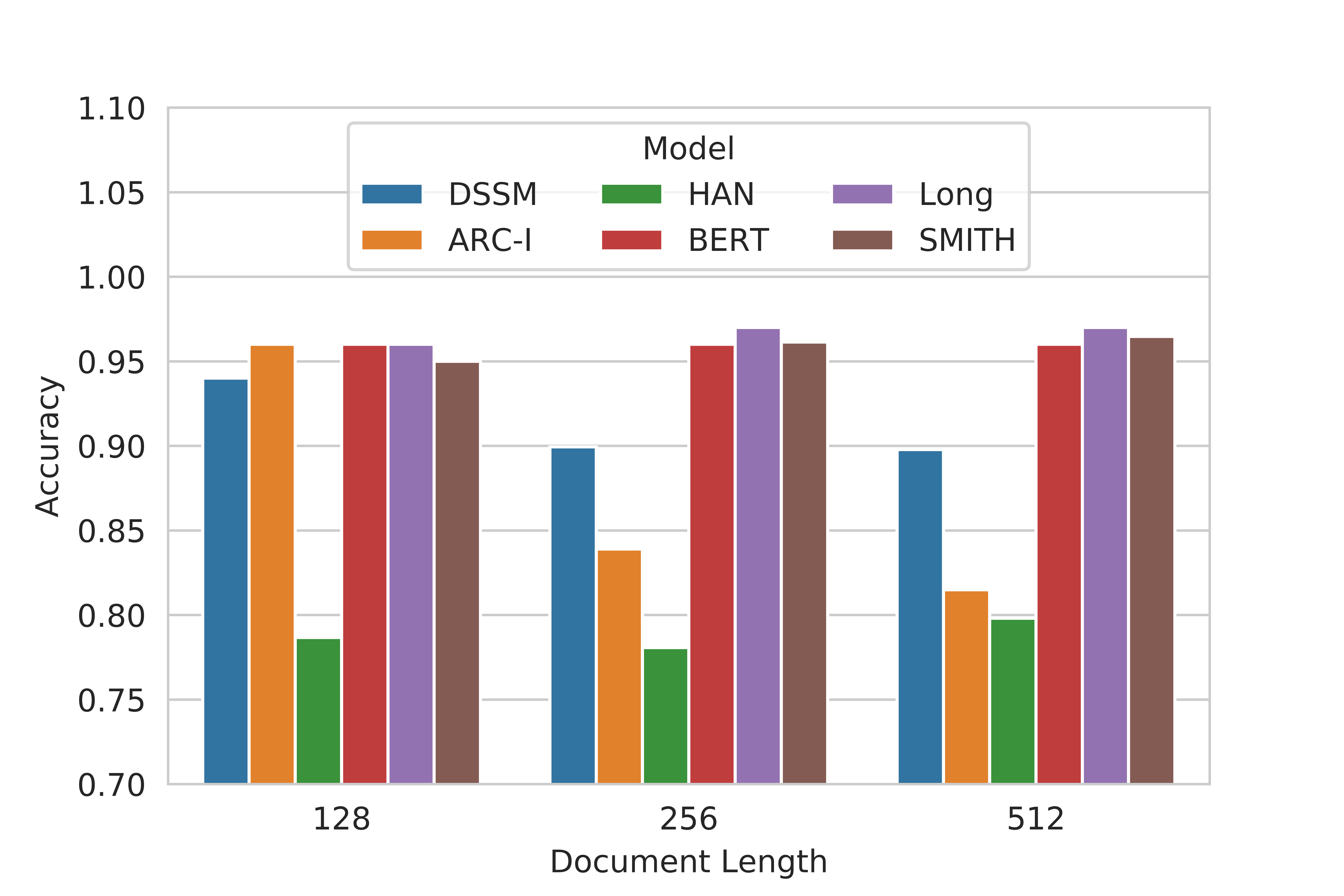}
    \caption{WIKI}
    \end{subfigure}
    \begin{subfigure}[]{0.33\linewidth}
    \includegraphics[width=\textwidth]{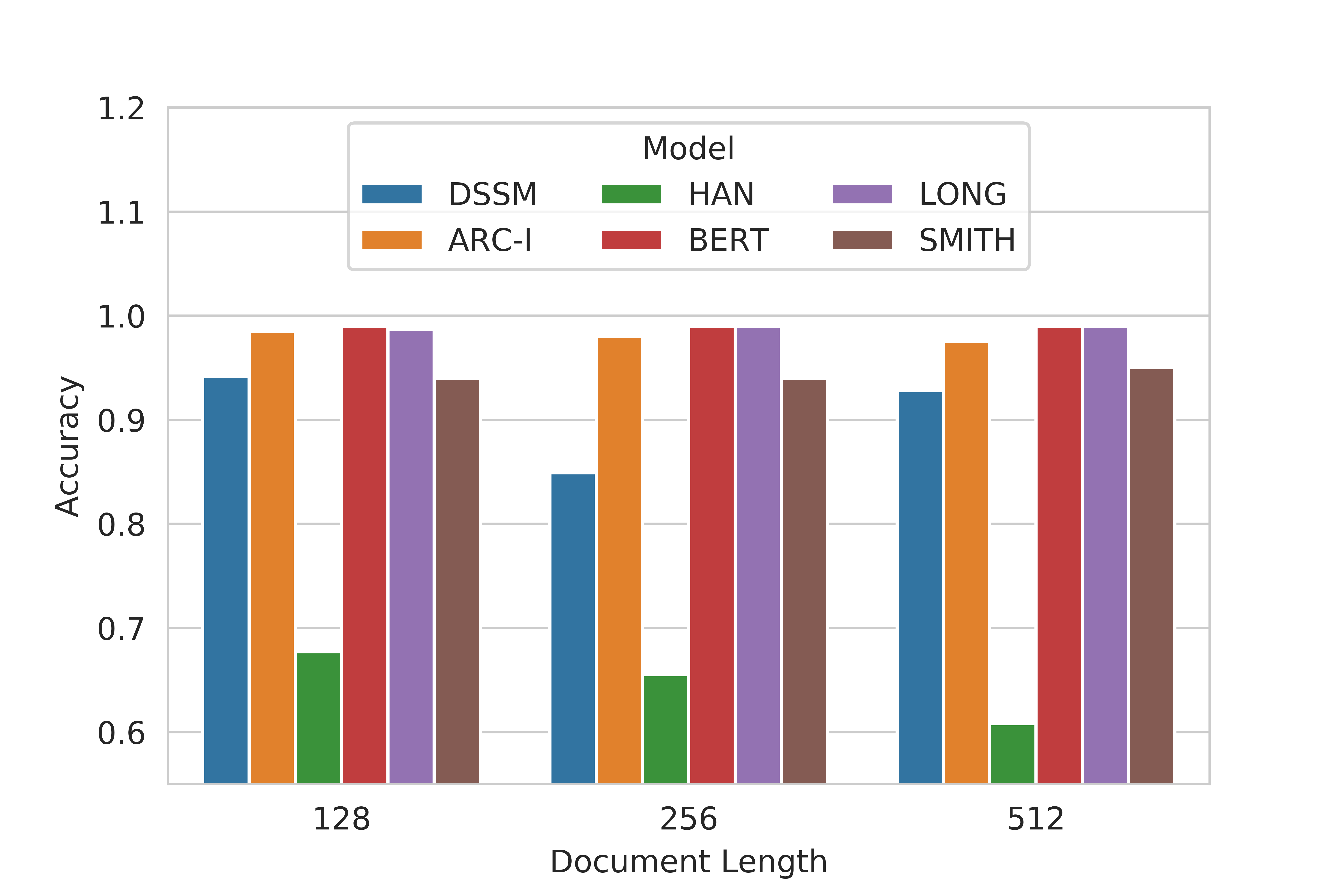}
    \caption{PAT}
    \end{subfigure}
    \caption{Document Length vs Accuracy upto 512 tokens}
    \label{fig_app:doclen_vs_accuracy_bert}
\end{figure*}

\begin{figure*}[h!]
\begin{subfigure}[]{0.33\linewidth}
    \includegraphics[width=\textwidth]{imgs/doclen_vs_acc/aan_doclen_vs_long.png}
    \caption{AAN}
    \end{subfigure}
    \begin{subfigure}[]{0.33\linewidth}
    \includegraphics[width=\textwidth]{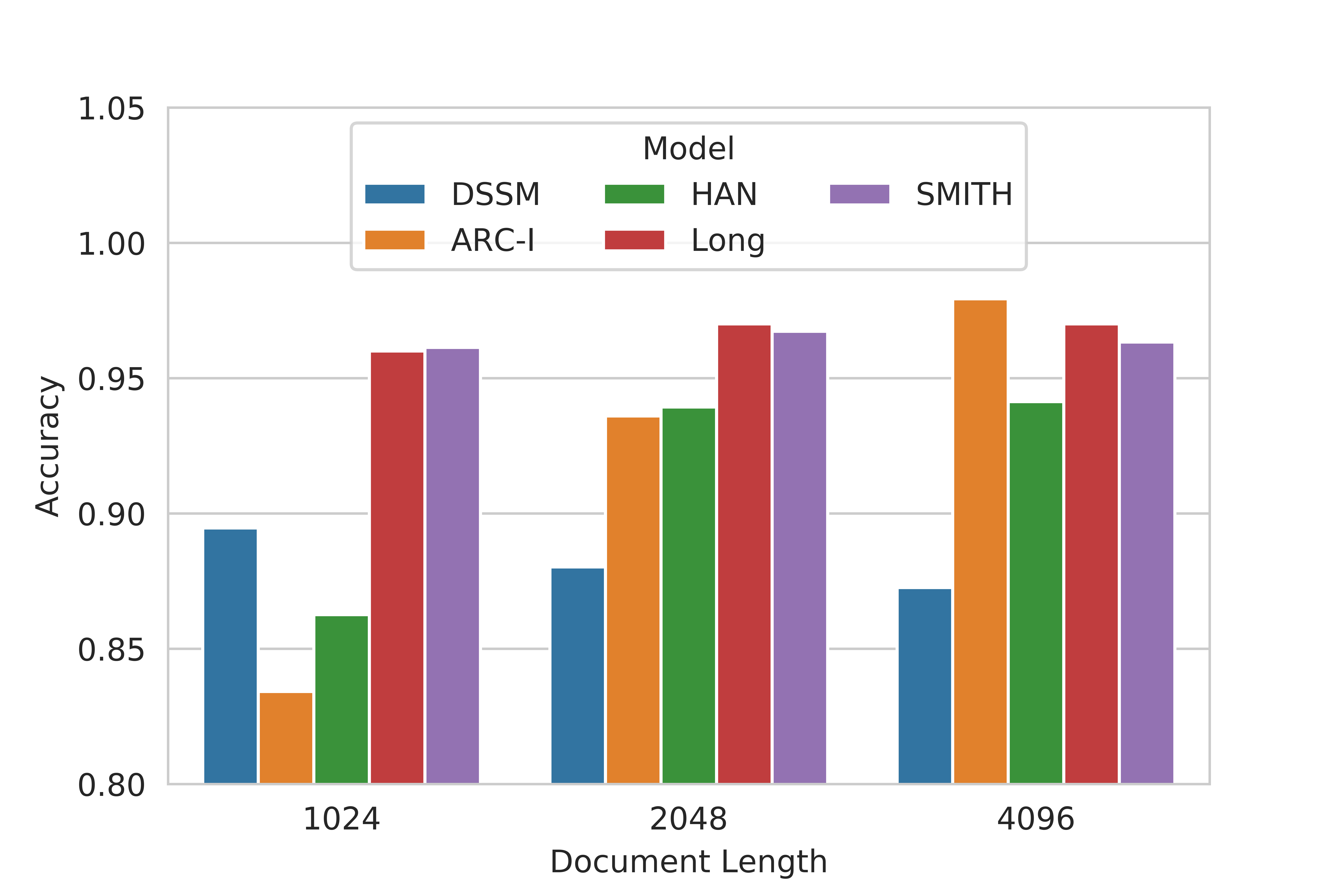}
    \caption{WIKI}
    \end{subfigure}
    \begin{subfigure}[]{0.33\linewidth}
    \includegraphics[width=\textwidth]{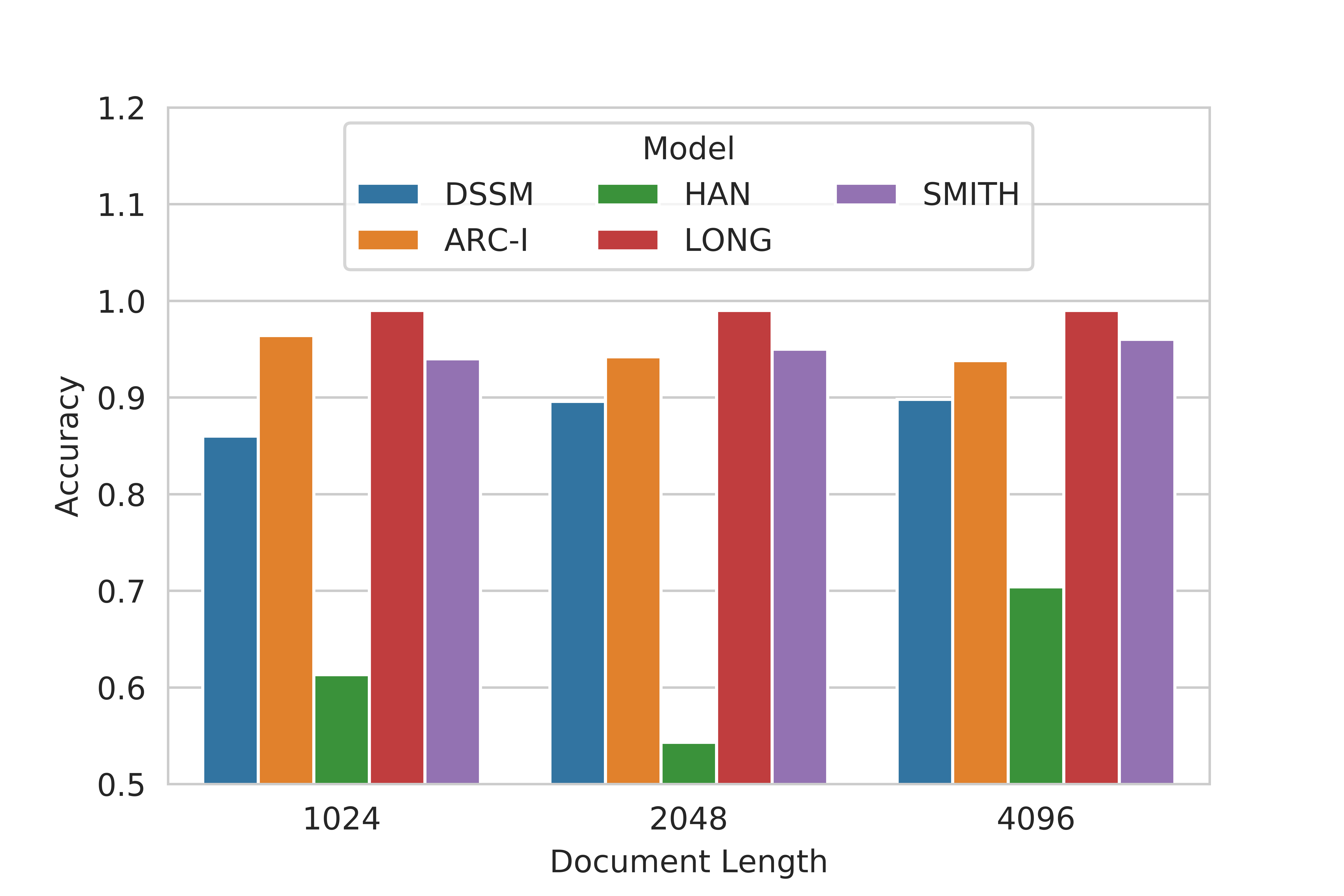}
    \caption{PAT}
    \end{subfigure}
    \caption{Document Length vs Accuracy upto 4096 tokens}
    \label{fig_app:doclen_vs_accuracy_long}
\end{figure*}

We check the robustness of simple models and transformer-based models for different document lengths on the task of document matching. From Figure~\ref{fig_app:doclen_vs_accuracy_bert} and Figure~\ref{fig_app:doclen_vs_accuracy_long}, we observe that the simple models DSSM and ARC-I, and the transformer-based models BERT and Longformer, though not specifically designed for long documents are robust for different document lengths. BERT can only handle up to 512 tokens at a time, and Longformer can only handle up to 4096 input tokens. HAN and SMITH, on the other hand, were specially designed for long documents and have a high variance in their performance on the document matching tasks for different document lengths.

We also experimented with longer documents ($>512$ tokens for BERT, and $>4096$ tokens for Longformer). We obtained the final document representation by dividing the document into chunks of their maximum allowed token length. We then aggregated these chunk representations. We experimented with the SUM and AVG aggregation techniques by taking the representations of the `[CLS]' token and `the pooler output' for these models. We observed an overall performance drop because of aggregation. The results were the same for both SUM and AVG aggregation techniques (Table ~\ref{tab_app:bert_cls_aggr}).

\subsection{Robustness to Text Perturbation} \label{sec_app:text_perturb}
\begin{figure*}[h!]
\begin{subfigure}[]{0.33\linewidth}
    \includegraphics[width=\textwidth]{imgs/perturb/aan_perturb.png}
    \caption{AAN}
    \end{subfigure}
    \begin{subfigure}[]{0.33\linewidth}
    \includegraphics[width=\textwidth]{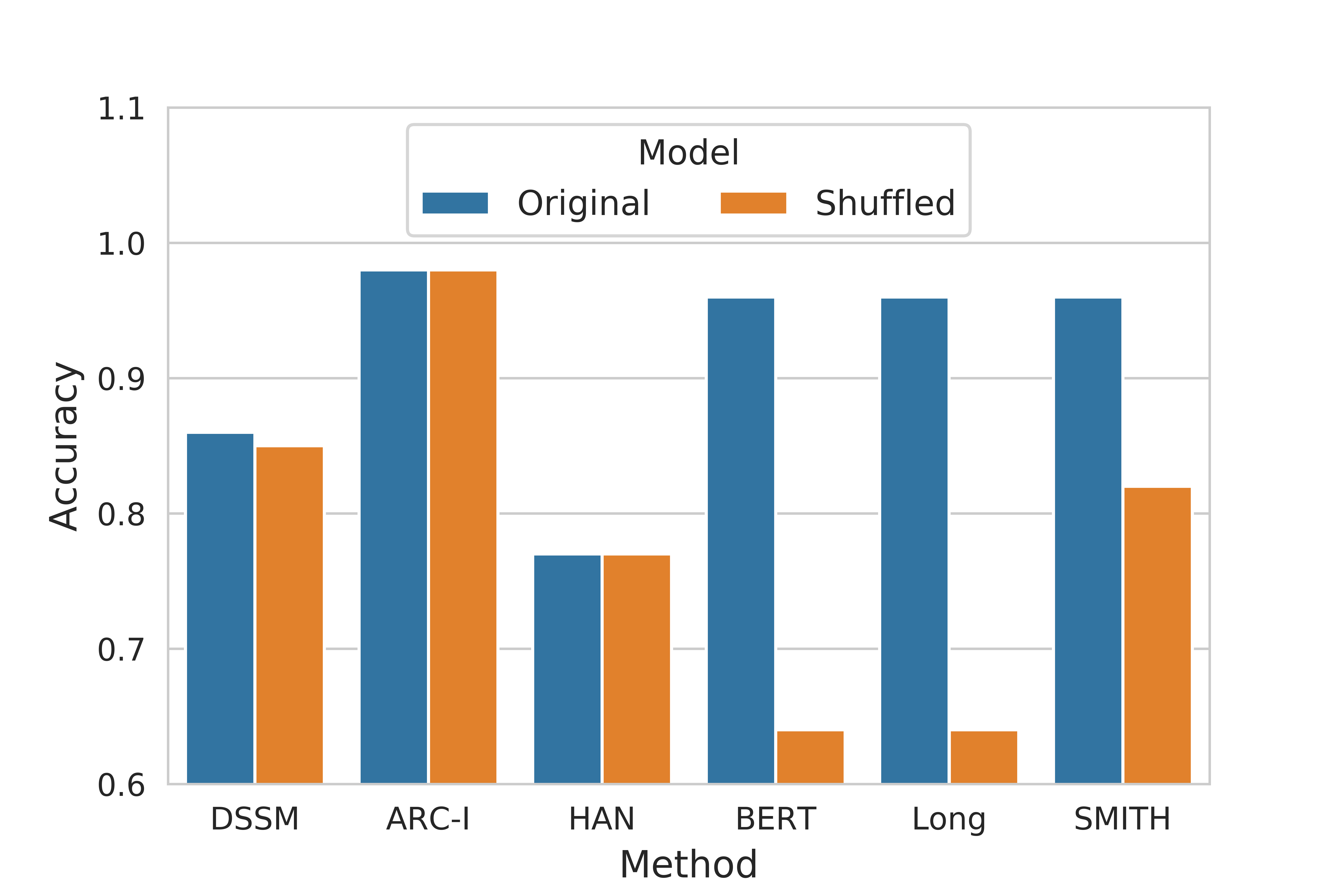}
    \caption{WIKI}
    \end{subfigure}
    \begin{subfigure}[]{0.33\linewidth}
    \includegraphics[width=\textwidth]{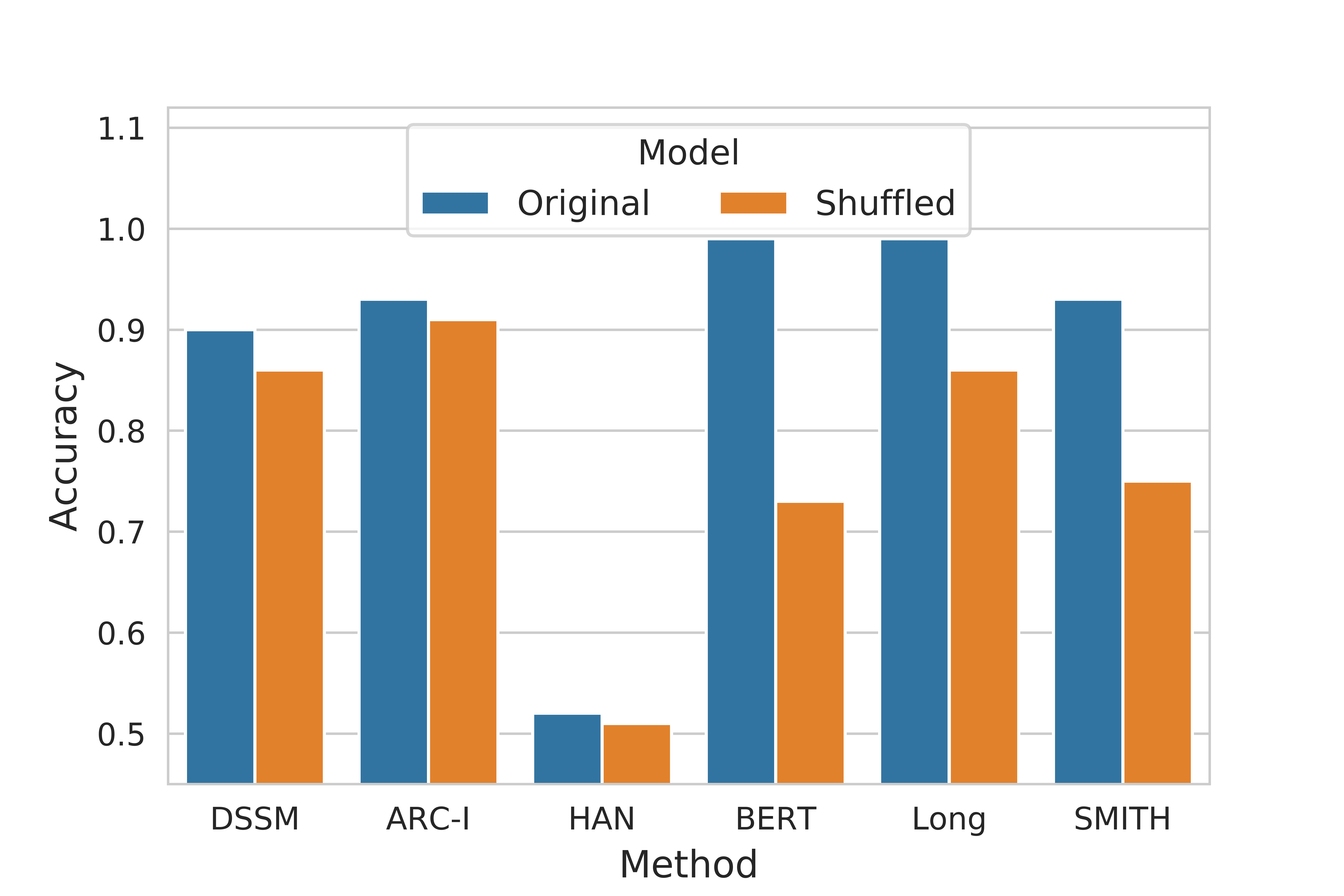}
    \caption{PAT}
    \end{subfigure}
    \caption{Original vs Shuffled Documents}
    \label{fig_app:acc_textperturb}
\end{figure*}

We randomly shuffle the documents before training different models and measure their test accuracy on the original document matching task (Figure~\ref{fig_app:acc_textperturb}) for all three datasets. {The first few paragraphs in Wikipedia articles, research papers, and patents are highly informative. We wanted to verify if the models give too much importance to the position of the initial text. In the context of long documents, just re-ordering the paragraphs of a document spanning pages should not have an effect on the downstream tasks of document matching. (Note: We use the term document matching broadly to refer to citation matching or document relevance. Given a document pair, we would like to verify if the two documents are relevant to each other.) In order to verify this assumption, we shuffle the paragraphs to distribute the important texts randomly and check the performance of all models on this downstream task. Although, we do observe a small drop after paragraph shuffling because the simple models do take into account a shallow context of the input text, the simple neural models overall prove more robust to text perturbation when compared to transformer-based models that take into account deep contextual information.}

\subsection{Best Input Embeddings for Simple Models}
\begin{table*}[h]
    \small
    \centering
    \begin{tabular}{|l|c|c|c|c|c|c|c|c|c|c|c|c|}
        \hline 
        & \multicolumn{4}{c|}{\textbf{AAN}} & \multicolumn{4}{c|}{\textbf{WIKI}} & \multicolumn{4}{c|}{\textbf{PAT}} \\
        \hline
        \textbf{Model} & \textbf{P} & \textbf{R} & \textbf{F1} & \textbf{Acc} & \textbf{P} & \textbf{R} & \textbf{F1}  & \textbf{Acc} & \textbf{P} & \textbf{R} & \textbf{F1} & \textbf{Acc}\\
        \hline
        DSSM-T & 0.768 & 0.809 & 0.787 & 0.780 & 0.823 & 0.939 & 0.877 & 0.869 & 0.869 & 0.957 & 0.911 & 0.905  \\
        ARC-I-T & 0.641 & 0.606 & 0.622 & 0.634 & 0.969 & 0.944 & 0.956 & 0.957 & 0.536 & 0.754 & 0.626 & 0.793 \\
        HAN-T & 0.665 & 0.885 & 0.759 & 0.720 & 0.911 & 0.929 & 0.920 & 0.920 & 0.477 & 0.857 & 0.618 & 0.751 \\
        
        \hline
        DSSM-G & 0.550 & 0.541 & 0.545 & 0.549 & 0.966 & \textbf{0.986} & 0.975 & 0.975 & \textbf{0.992} & \textbf{0.998} & \textbf{0.995} & \textbf{0.995} \\
        ARC-I-G & 0.643 & 0.872 & 0.734 & 0.676 & \textbf{0.992} & {0.983} & \textbf{0.987} & \textbf{0.987} & 0.905 & 0.963 & 0.933 & 0.929 \\
        HAN-G & 0.504 & 0.881 & 0.641 & 0.507 & 0.935 & 0.984 & 0.959 & 0.958 & 0.609 & 0.848 & 0.709 & 0.522 \\
        
        \hline
        DSSM-D & \textbf{0.852} & 0.763 & \textbf{0.805} & \textbf{0.815} & 0.933 & 0.984 & 0.958 & 0.957 & 0.841 & 0.959 & 0.896 & 0.949 \\
        ARC-I-D & 0.841 & 0.763 & 0.800 & 0.809 & 0.987 & 0.985 & 0.986 & 0.986 & 0.967 & 0.958 & 0.962 & 0.983 \\
        HAN-D & 0.709 & \textbf{0.919} & 0.801 & 0.771 & 0.875 & 0.859 & 0.866 & 0.873 & 0.946 & 0.996 & 0.970 & 0.975 \\
        \hline
    \end{tabular}
     \caption{Comparison of different input aggregation techniques: (i) charTrigrams (T), (ii) GloVe Embeddings (G), and (iii) Doc2Vec embeddings (D), for simple models.}
    \label{tab_app:input_aggr}
\end{table*}

\begin{table*}[!h]
    \small
    \centering
    \begin{tabular}{|l|c|c|c|c|c|c|c|c|c|c|c|c|}
        \hline 
        & \multicolumn{4}{c|}{\textbf{AAN}} & \multicolumn{4}{c|}{\textbf{WIKI}} & \multicolumn{4}{c|}{\textbf{PAT}} \\
        \hline
        \textbf{Model} & \textbf{P} & \textbf{R} & \textbf{F1} & \textbf{Acc} & \textbf{P} & \textbf{R} & \textbf{F1}  & \textbf{Acc} & \textbf{P} & \textbf{R} & \textbf{F1} & \textbf{Acc}\\
        \hline
        BERT-CLS & 0.992 & 0.579 & 0.732 & 0.637 & 0.998 & 0.499 & 0.666 & 0.499 & 1 & 0.639 & 0.783 & 0.714 \\
        BERT-POOL & 0.572 & 0.992 & 0.726 & 0.625 & 1 & 0.500 & 0.667 & 0.501 & 1 & 0.637 & 0.778 & 0.711 \\
        \hline
        LONG-CLS & 0.599 & 0.842 & 0.699 & 0.743 & 0.968 & 0.764 & 0.854 & 0.835 & 0.927 & 0.911 & 0.919 & 0.917 \\
        LONG-POOL & 0.727 & 0.819 & 0.770 & 0.783 & 0.994 & 0.812 & 0.894 & 0.883 & 0.996 & 0.918 & 0.955 & 0.953 \\
        \hline
    \end{tabular}
    \caption{Aggregation using the [CLS] token, and the pooler output [POOL] from BERT and Longformer for documents $>512$ and $>4096$ tokens for BERT and Longformer, respectively. The results were the same for SUM and AVG aggregation techniques.}
    \label{tab_app:bert_cls_aggr}
\end{table*}

For simple models, we evaluate if the input vector representations play a role in the final results. We use the following input representations.

\begin{itemize}
    \item \textbf{Tri-Gram Hashing (T)}: Bag-of-charTrigrams is a technique for word hashing \cite{huang2013learning} where each word is broken down into character trigrams (charTrigrams). Since, the number of possible charTrigarms are fixed and limited, this is a scalable solution for long documents. The charTrigrams are obtained for every token in the input text after appending the symbol ‘\#’ before and after every token. For example, the word `good' [\#good\#] is split into [\#go, goo, ood, od\#] and then mapped to a 30,621 dimensional hash table. This vector representation for the document is then given as input to the models. {For DSSM, each document is represented as a bag-of-charTrigrams and given as input to the model. For ARC-I and HAN, we split each document into $n$ chunks which are represented in the form of a trigram hash. We construct a matrix of size $n~\times~\text{trigram hash}$ for the entire document which is given input to ARC-I and HAN.}
    
    \item \textbf{GloVe Embeddings (G)}:
    GloVe \cite{pennington2014glove} is an unsupervised learning algorithm for obtaining vector representations for words. We download the pre-trained GloVe embeddings and get the vector representations for words in a long document. These vector representations are given as input to different models. {For DSSM, we divide the document into chunks of a specified maximum length. We then take GloVe embedding representation of tokens for each chunk upto a maximum length and average them to get a document representation. For ARC-I and HAN, each document is represented as a matrix of size $[\text{embedding dimension}~\times~\text{max length}]$ in each document.}

    \item \textbf{Doc2Vec Embeddings (D)}: Doc2Vec \cite{le2014distributed} embeddings can be used to get vector representations for a document. We train Doc2Vec models from scratch on different datasets to get relevant document representations. These document representations are then given as input to different document matching models. 
\end{itemize}

Table~\ref{tab_app:input_aggr} shows the model performance of the simple models for the above three input representations. We observe that using GloVe (G) and Doc2Vec (D) input embeddings improve the model performance of the simple models overall. 

\subsection{Different Aggregation Techniques for Transformer Based Models} \label{app:trans_aggr}

We experiment with different aggregation techniques (SUM and AVG) for $>512$ tokens for BERT, and $>4096$ tokens for Longformer. We chunk the documents and aggregate the representations from the `[CLS]' token and `the pooler output'. The results can be seen in Table~\ref{tab_app:bert_cls_aggr} and were the same for SUM and AVG aggregation techniques. Aggregation resulted in an overall performance drop when compared to just truncating the documents up to 512 tokens and 4096 tokens for BERT and Longformers, respectively for the document matching.

\end{document}